\documentclass{article}

\usepackage[final]{neurips_2024}

\usepackage[utf8]{inputenc} 
\usepackage[T1]{fontenc}    
\usepackage{hyperref}       
\usepackage{url}            
\usepackage{booktabs}       
\usepackage{amsfonts}       
\usepackage{nicefrac}       
\usepackage{microtype}      
\usepackage{xcolor}         

\newtheorem{theorem}{\bf{Theorem}}
\newtheorem{lemma}{\bf{Lemma}}
\newtheorem{proposition}{\bf{Proposition}}

\newtheorem{definition}{\bf{Definition}}

\newtheorem{example}{\bf{Example}}
\def\QED{~\rule[-1pt]{5pt}{5pt}\par\medskip}
\newenvironment{proof}{\emph{Proof.}}{\hfill\QED}

\newenvironment{theorem*}[1][]{%
    \par\addvspace{\topsep}
    \noindent{\bfseries Theorem\ifx&#1&\relax\else\ (#1)\fi\quad}\itshape
    \ignorespaces
}{%
    \par\addvspace{\topsep}
}

\newenvironment{lemma*}[1][]{%
    \par\addvspace{\topsep}
    \noindent{\bfseries Lemma\ifx&#1&\relax\else\ (#1)\fi\quad}\itshape
    \ignorespaces
}{%
    \par\addvspace{\topsep}
}

\newenvironment{proposition*}[1][]{%
    \par\addvspace{\topsep}
    \noindent{\bfseries Proposition\ifx&#1&\relax\else\ (#1)\fi\quad}\itshape
    \ignorespaces
}{%
    \par\addvspace{\topsep}
}

\newenvironment{remark*}[1][]{%
    \par\addvspace{\topsep}
    \noindent{\bfseries Remark\ifx&#1&\relax\else\ (#1)\fi\quad}\itshape
    \ignorespaces
}{%
    \par\addvspace{\topsep}
}

\DeclareMathAlphabet{\altmathcal}{OMS}{cmsy}{m}{n} 

\newcommand{\argmin}[1]{\underset{#1}{\operatorname{arg \ min \ }}}
\newcommand{\argmax}[1]{\underset{#1}{\operatorname{arg \ max \ }}}

\newcommand{\KL}[2]{\operatorname{\mathcal{D}_{KL}}\left(#1 \, \middle\| \, #2\right)}

\newcommand{\xmath}[1] {\ensuremath{#1}\xspace}
\newcommand{\blmath}[1] {\xmath{\bm{#1}}}

\newcommand{\B}{\blmath{B}}

\newcommand{\E}{\blmath{E}}

\newcommand{\M}{\blmath{M}}

\newcommand{\T}{\blmath{T}}

\newcommand{\vb}{\blmath{b}}
\newcommand{\ve}{\blmath{e}}

\newcommand{\vv}{\blmath{v}}

\newcommand{\veta}{\blmath{\eta}}

\newcommand{\1}{\blmath{1}}

\newcommand{\mc}{\mathcal}

\newcommand{\bb}{\mathbb}

\newcommand{\indicator}[1]{\mathbbm 1_{#1}}

\newcommand{\abs}[1]{\left| #1 \right|}

\newcommand{\prob}[1]{\bb P\left( #1 \right)}
\newcommand{\expect}[2]{\bb E_{#1}\left[ #2 \right]}

\newcommand{\half}{\mbox{$\frac12$}}

\renewcommand{\mathbf}{\boldsymbol}

\usepackage{amsmath}
\usepackage{enumitem}
\usepackage{multirow}
\usepackage{tablefootnote}
\usepackage{bm}
\usepackage{bbm}
\usepackage{amssymb} 
\usepackage{xspace}
\usepackage{tikz} 
\usepackage[colorinlistoftodos]{todonotes} 
\usepackage{pgfplots} 
\usepgfplotslibrary{groupplots}
\pgfplotsset{compat=1.17}
\usepackage{graphicx}
\usepackage{subcaption}
\usepackage{graphicx}
\usepackage{cleveref}
\usepackage{float}
\usepackage{wrapfig}

\newcommand{\tveta}{\widetilde{\veta}}

\title{
    Label Noise: Ignorance Is Bliss
    }

\author{%
  Yilun Zhu
    \\
    EECS\\
    University of Michigan\\
  \texttt{allanzhu@umich.edu} \\
  \And
  Jianxin Zhang \\
  EECS\\
  University of Michigan\\
  \texttt{jianxinz@umich.edu} \\
  \AND 
  Aditya Gangrade  \\
  ECE \\
  Boston University\\
  \texttt{gangrade@bu.edu
  } \\
  \And
  Clayton Scott \\
  EECS, Statistics \\
  University of Michigan\\
  \texttt{clayscot@umich.edu} \\
}

\begin{document}

\maketitle

\begin{abstract}
    We establish a new theoretical framework for learning under multi-class, instance-dependent label noise. 
    This framework casts learning with label
    noise as a form of domain adaptation, in particular, domain adaptation
    under posterior drift. 
    We introduce the concept of \emph{relative signal strength} (RSS), a pointwise measure that quantifies the transferability from noisy to clean posterior. 
    Using RSS, we establish nearly matching upper and lower bounds on the excess risk. 
    Our theoretical findings support 
    the simple
    \emph{Noise Ignorant Empirical Risk Minimization (NI-ERM)} principle,
    which minimizes empirical risk while ignoring label noise.
    Finally, we translate this theoretical insight into practice: by
    using NI-ERM to fit a linear classifier on top of a self-supervised
    feature extractor, we achieve state-of-the-art performance on the
    CIFAR-N data challenge.

\end{abstract}

\section{Introduction}

The problem of classification with label noise can be stated in terms of random variables $X$, $Y$, and $\widetilde{Y}$, where $X$ is the feature vector, $Y \in \{1,\ldots, K\}$ is the true label associated to $X$, and $\widetilde{Y} \in \{1,\ldots, K\}$ is a noisy version of $Y$. The learner has access to i.i.d. realizations of $(X,\widetilde{Y})$, and the objective is to learn a classifier that optimizes the risk associated with $(X,Y)$. 

In recent years, there has been a surge of interest in the challenging setting of instance (i.e., feature) dependent label noise, in which $\widetilde{Y}$ can depend on both $Y$ and $X$. While several algorithms have been developed, there remains relatively little theory regarding algorithm performance and the fundamental limits of this learning paradigm.

This work develops a theoretical framework for learning under multi-class, instance-dependent label noise. Our framework hinges on the concept of \emph{relative signal strength}, which is a point-wise measure of ``noisiness'' in a label noise problem. Using relative signal strength to charachterize the difficulty of a label noise problem, we establish nearly matching upper and lower bounds for excess risk. 
We further identify distributional assumptions that ensure that the lower and upper bounds tend to zero as  the sample size $n$ grows, implying that consistent learning is possible.

Surprisingly, \emph{Noise Ignorant Empirical Risk Minimization (NI-ERM)} principle, which conducts empirical risk minimization as if no label noise exists, 
is (nearly) minimax optimal. 
To translate this insight into practice, we use NI-ERM to fit a linear classifier on top of a self-supervised feature extractor, achieving state-of-the-art performance on the CIFAR-N data challenge.



\section{Literature review} 

Theory and algorithms for classification with label noise 
are often based on different probabilistic models. Such models may be categorized according on how $\widetilde{Y}$ depends on $Y$ and $X$. The simplest model is symmetric noise, where the distribution of $\widetilde{Y}$ is independent of $Y$ and $X$ \citep{angluin1988learning}. In this case, the probability that $\widetilde{Y} =k$ is the same for all $k \ne Y$, regardless of $Y$ and $X$. In this setting, it is easy to show that minimizing the noisy excess risk (associated to the 0/1 loss) implies minimizing the clean excess risk, a property known as \emph{immunity}. When immunity holds, there is no need to modify the learning algorithm on account of noisy labels. In other words, the learner may be \emph{ignorant} of the label noise and still learn consistently.

A more general model is classification with label dependent noise, in which the distribution of $\widetilde{Y}$ depends on $Y$, but not $X$. 
Many practical algorithms have been developed over the years, based on principles including data re-weighting \citep{liu2015classification}, robust training \citep{han2018coteaching, liu2020early, foret2021sharpnessaware, liu2022robust} and data cleaning \citep{brodley1999identifying, northcutt2021confident}.
Consistent learning algorithms still exist, such as those based on loss correction \citep{natarajan2013learning, patrini2017making, van2018theory, liu2020peer, zhang2022learning}. 
These approaches assume knowledge of the noise transition probabilities, which can be estimated under some identifiability assumptions \citep{scott2013classification, zhang2021learning}.


In the most general setting, that of instance dependent label noise, the distribution of $\widetilde{Y}$ depends on both $Y$ and $X$. 
While algorithms are emerging \citep{cheng2021learning, zhu2021second, wang2022estimating, yang2023parametrical}, theory has primarily focused on the binary setting. \citet{scott2019generalized} establishes immunity for a Neyman-Pearson-like performance criterion under a \emph{posterior drift} model, discussed in more detail below. \citet{cannings2020classification} 
establish an upper bound for excess risk under the strong assumption that the optimal classifiers for the clean and noisy distributions are the same. 

Closest to our work, \cite{im2023binary} derive excess risk upper and lower bounds, and reach a similar conclusion, that noise-ignorant ERM attains the lower bound up to a constant factor. 
Our results, based on the new concept of relative signal strength, provide a more refined analysis. 

Additional connections between our contributions and prior work are made throughout the paper.

\section{Problem statement 
}
\label{sec:setup}


\emph{Notation.}  \(\mathcal{X}\) denotes the feature space and \(\mathcal{Y} = \{1,2, \dots, K\} \) denotes the label space, with \(K \in \mathbb{N}\). The \(K\)-simplex is \(\Delta^K := \{ p \in \mathbb{R}^K : \forall i, p_i \geq 0, \sum p_i = 1\}\). A $K \times K$ matrix is \emph{row stochastic} if all of its rows are in $\Delta^K$. Denote the $i$-th element of a vector $\vv$ as $[\vv]_i,$ and the $(i,j)$-th element of a matrix $\M$ as $[\M]_{i, j}.$


In conventional multiclass classification, we observe training data $(X_1, Y_1), \dots, (X_n, Y_n)$ drawn i.i.d. from a joint distribution $P_{XY}$. The marginal distribution of $X$ is denoted by $P_X$, and the \emph{class posterior} probabilities 
$P_{Y|X=x}$ are captured by a $K$-simplex-valued vector $\veta: \mathcal{X} \rightarrow \Delta^K $, where the $j$-th component of the vector is $[\veta(x)]_j = \prob{Y = j \mid X = x}$. A classifier $f: \mc{X} \rightarrow \mc{Y}$ maps an instance $x$ to a class $f(x) \in \mathcal{Y}$. 
Denote the risk of a classifier $f$ with respect to distribution $P_{XY}$ as $R(f) = \expect{(X, Y) \sim P_{XY}}{ \indicator{ \left\{ f(X) \neq Y \right\} } }$. The Bayes optimal classifier for $P_{XY}$ is $f^*(x) \in \arg \max \veta(x)$. The Bayes risk, which is the minimum achievable risk, is denoted as $R^* = R(f^*) = \inf_f R(f)$.

We consider the setting where, instead of the true class label $Y$, a noisy label $\widetilde{Y}$ is observed. The training data $(X_1, \widetilde{Y}_1), \dots, (X_n, \widetilde{Y}_n)$ can be viewed as an i.i.d. sample drawn from a ``noisy'' distribution $P_{X \widetilde{Y}}$. We define $P_{\widetilde{Y}|X=x}$, $\widetilde{\veta}$, $\widetilde{R}$ and $\tilde{f}^*$ analogously to the ``clean'' distribution $P_{XY}$.

The goal of learning from label noise is to find a classifier that is able to minimize the ``clean test error,'' that is, the risk $R$ defined w.r.t. $P_{XY}$, even though the learner has access to only corrupted training data \((X_i, \widetilde{Y}_i) \stackrel{\text{i.i.d.}}{\sim} P_{X \widetilde{Y}}\).

\paragraph{Noise transition perspective.}
Traditionally, label noise is modeled through the joint distribution of $(X, Y, \widetilde{Y})$. This joint distribution is governed by $P_X$, the clean class posterior $P_{Y|X}$, and a matrix-valued function 
\[
\E: \mathcal{X} \to \{\M \in \mathbb{R}^{K \times K} : \M \text{ is row stochastic}\},
\]
known as the \emph{noise transition matrix}. The $(i, j)$-th entry of the matrix is defined as:
\[
[\E(x)]_{i, j} = \prob{\widetilde{Y} = j \mid Y = i, X = x }.
\]
This implies that the noisy and clean class posteriors are related by $\widetilde{\veta}(x) = \E(x)^\top \veta(x)$, where $^{\text{$\top$}}$ denotes the matrix transpose.

\paragraph{Domain adaptation perspective.}
Alternatively, label noise learning can be framed as a domain adaptation problem. In this view, $P_{X\widetilde{Y}}$ represents the source domain, and $P_{XY}$ represents the target domain. The relationship between the two domains is characterized by ``posterior drift,'' meaning that while the source and target share the same $X$-marginal, the class posteriors (i.e., the distribution of labels given $X$) may differ \citep{scott2019generalized, cai2021transfer, maity2021linear}. Thus, a label noise problem can also be described by a triple $(P_X, \veta, \widetilde{\veta})$. 

The two perspectives are equivalent, as discussed in Appendix \ref{sec:equivalentperspective}. In this work, we emphasize the domain adaptation perspective for Sections \ref{sec:rss} and \ref{sec:bounds}, and the noise transition perspective for Section \ref{sec:immunity}.

\section{Relative signal strength}
\label{sec:rss}



To study label noise, we introduce the concept of \emph{relative signal strength} (RSS). This is a pointwise measure of how much ``signal'' (certainty about the label) is contained in the noisy distribution relative to the clean distribution. Previous work \citep{cannings2020classification, cai2021transfer} has examined a related concept within the context of binary classification, under the restriction that clean and noisy Bayes classifiers are identical. Our definition incorporates multi-class classification and relaxes the requirement that the clean and noisy Bayes classifiers agree.






\begin{definition}[Relative Signal Strength]
    For any class probability vectors $\veta, \widetilde{\veta} $, define the \emph{relative signal strength} (RSS) at $x \in \mc{X}$ as 
    \begin{equation}
    \label{eqn:rss}
        \mc{M}(x; \veta, \widetilde{\veta}) = \min_{j \in \mc{Y}} \ \ \frac{ \max_i [\widetilde{\veta}(x)]_i - [\widetilde{\veta}(x)]_{j}  }{ \max_i [\veta(x)]_i - [\veta(x)]_{j} },
    \end{equation}
    where $0/0 := + \infty$. Furthermore, for $\kappa \in [0, \infty)$, denote the set of points whose RSS exceeds $\kappa$ as 
    \begin{align*}
        \mc{A}_{\kappa} (\veta, \widetilde{\veta}) = \left\{ x \in \mc{X}: \mc{M}(x; \veta, \widetilde{\veta}) > \kappa \right\}.
    \end{align*}
    \label{def:rss}
\end{definition}

$\mc{M}(x; \veta, \widetilde{\veta})$ is a point-wise measure of how much ``signal'' the noisy posterior contains about the clean posterior. 
To gain some intuition, first notice that if the noisy Bayes classifier predicts a different class than the clean Bayes classifier, the RSS is 0 by taking $j =  \arg \max \widetilde{\veta}$ 
(assuming for simplicity that the $\arg \max$ is a singleton set). Now suppose the clean and noisy Bayes classifiers \emph{do} make the same prediction at $x$, say $i^*$, and consider a fixed $j$. 
If 
\[
\frac{[\widetilde{\veta}(x)]_{i^*} - [\widetilde{\veta}(x)]_{j}}{[{\veta}(x)]_{i^*} - [{\veta}(x)]_{j}}
\]
is small, it means that the clean Bayes classifier is relatively certain that $j$ is not the correct clean label, while the noisy Bayes classifier is less certain that $j$ is not the correct noisy label. Taking the minimum over $j$ gives the relative signal strength at $x$. As we formalize in the next section, a large RSS at $x$ ensures that a small (pointwise) \emph{noisy} excess risk at $x$ implies a small (pointwise) \emph{clean} excess risk. 
To gain more intuition, consider the following examples.





\begin{example}
When $\veta (x) = [0 \ \ 1 \ \ 0 ]^\top$ and $\widetilde{\veta} (x) = [0.3 \ \ 0.6 \ \ 0.1]^\top$,
\begin{align*}
    \mc{M}(x; \veta, \widetilde{\veta}) 
    = \min_{j \in \mc{Y}} \ \ \frac{ \max_i [\widetilde{\veta}(x)]_i - [\widetilde{\veta}(x)]_{j}  }{ \max_i [\veta(x)]_i - [\veta(x)]_{j} } 
    = \frac{  [\widetilde{\veta}(x)]_2 - [\widetilde{\veta}(x)]_{1}  }{  [\veta(x)]_2 - [\veta(x)]_{1} } = \frac{0.6 - 0.3}{1  - 0} = 0.3.
\end{align*}
Here, first of all, $\arg \max \veta = \arg \max \widetilde{\veta} = 2$, i.e., the clean and noisy Bayes classifier give the same prediction.
What's more, $\mc{M}(x; \veta, \widetilde{\veta}) < 1$ because
the clean Bayes classifier is absolutely certain about its prediction, while the noisy Bayes classifier is much less certain.
\end{example}

\begin{example}
When $\veta (x) = [0 \ \ 1 \ \ 0 ]^\top$ and $\widetilde{\veta} (x) = [0 \ \ 0 \ \ 1 ]^\top$,
\begin{align*}
    \mc{M}(x; \veta, \widetilde{\veta}) 
    = \min_{j \in \mc{Y}} \ \ \frac{ \max_i [\widetilde{\veta}(x)]_i - [\widetilde{\veta}(x)]_{j}  }{ \max_i [\veta(x)]_i - [\veta(x)]_{j} } 
    = \frac{  [\widetilde{\veta}(x)]_3 - [\widetilde{\veta}(x)]_{3}  }{  [\veta(x)]_2 - [\veta(x)]_{3} } = \frac{1 - 1}{1  - 0} = 0.
\end{align*}
The  zero signal strength results from $\widetilde{\veta}$ and $\veta$ leading to different predictions about $\arg \max$. 
\end{example}

\begin{example}[Comparison to KL divergence]
    When $\veta (x) = [0.05 \ \ 0.7 \ \ 0.25 ]^\top$, and $\widetilde{\veta}^{(1)} (x) = [0.25 \ \ 0.7 \ \ 0.05 ]^\top, \ \ \widetilde{\veta}^{(2)} (x) = [0.1 \ \ 0.6 \ \ 0.3 ]^\top$,
    \begin{align*}
        \frac{1}{\KL{\veta}{\widetilde{\veta}^{(1)}}} < \frac{1}{\KL{\veta}{\widetilde{\veta}^{(2)}}} \quad \text{while} \quad
        \mc{M} \left(x; \veta, \widetilde{\veta}^{(1)} \right) > \mc{M} \left( x; \veta, \widetilde{\veta}^{(2)} \right).
    \end{align*}

    Here, $\widetilde{\veta}^{(2)}$ is ``closer'' to $\veta$ in terms of KL divergence, but $\widetilde{\veta}^{(1)}$ provides more information in terms of predicting the $\arg \max$ of $\veta$. There is no conflict: KL divergence considers the similarity between two (whole) distributions, while the task of classification only focuses on predicting the $\arg \max$. 
    
    This also illustrates why our notion of RSS is better suited for the label noise problem than other general-purpose distance measures between distributions.
    \label{eg:KL_RSS}
\end{example}

A desirable learning scenario would be if $\mc{A}_{\kappa} (\veta, \widetilde{\veta}) = \mc{X}$ for some large $\kappa$, indicating that the signal strength is big across the entire space. Unfortunately, this ideal situation is generally not achievable. To gain some insight, consider the following result, proved in Appendix \ref{subsec:signal_subset_proof}.


\begin{proposition}
    $\mc{A}_{0} (\veta, \widetilde{\veta}) = \big\{x \in \mc{X}: \argmax{} \widetilde{\veta}(x) \subseteq \argmax{} {\veta}(x) \big\}.$
    \label{prop:signal_subset}
\end{proposition}
If we assume that both $\arg \max$ sets are singletons, this result indicates that $\mc{A}_{0}$, the region with positive RSS, is the region where the true and noisy Bayes classifiers agree. Accordingly, $\mc{X} \setminus \mc{A}_{0}$, the zero signal region, is the region where the clean and noisy Bayes decision rules differ. 
The ``region of strong signal,'' $\mc{A}_{\kappa}$, is a subset of $\mc{A}_{0}$. Since the clean and noisy Bayes classifiers will typically disagree for at least some $x$, $\mc{A}_0 \neq \mc{X}$ in general. We note that the strong assumption that $\mc{A}_0 = \mc{X}$ has been made in prior studies \citep{cannings2020classification, cai2021transfer}. Our notion of RSS relaxes this assumption
and provides a unified view.

\subsection{RSS in binary classification}
We can express relative signal strength more explicitly in the binary setup. Let $\eta(x) := [\veta(x)]_1 = \prob{Y = 1 \mid X = x}$ and $\widetilde{\eta}(x) := [\widetilde{\veta}(x)]_1 = \bb{P}\big( \widetilde{Y} = 1 \mid X = x \big)$. In standard binary classification, the \emph{margin} \citep{tsybakov2004optimal, massart2006risk}, defined as $\abs{ \eta(x) - \half }$, serves as a pointwise measure of signal strength. Our notion of relative signal strength (RSS) can be interpreted as an extension of this concept in the context of label noise learning.

\begin{proposition}
    In the binary setting, for $\kappa \geq 0$,
    \begin{align*}
        \mc{M}(x; \eta, \widetilde{\eta}) = \max{ \left\{ \frac{  \widetilde{\eta}(x) - \half }{ \eta(x) - \half } , \ 0 \right\} }, \qquad \text{and} \qquad
        \mc{A}_{\kappa} (\eta, \widetilde{\eta})  = \left\{ x \in \mc{X}: \frac{  \widetilde{\eta}(x) - \half }{ \eta(x) - \half } > \kappa \right\}.
    \end{align*}
    \label{prop:rss_binary}
\end{proposition}

In other words, RSS can be viewed as a ``relative'' margin.

\begin{example}
Illustration of relative signal strength in a binary classification setup (Figure \ref{fig:relative_signal_strength_example}). 
\begin{figure}[h]
    \centering
    \includegraphics[width=0.9\linewidth]{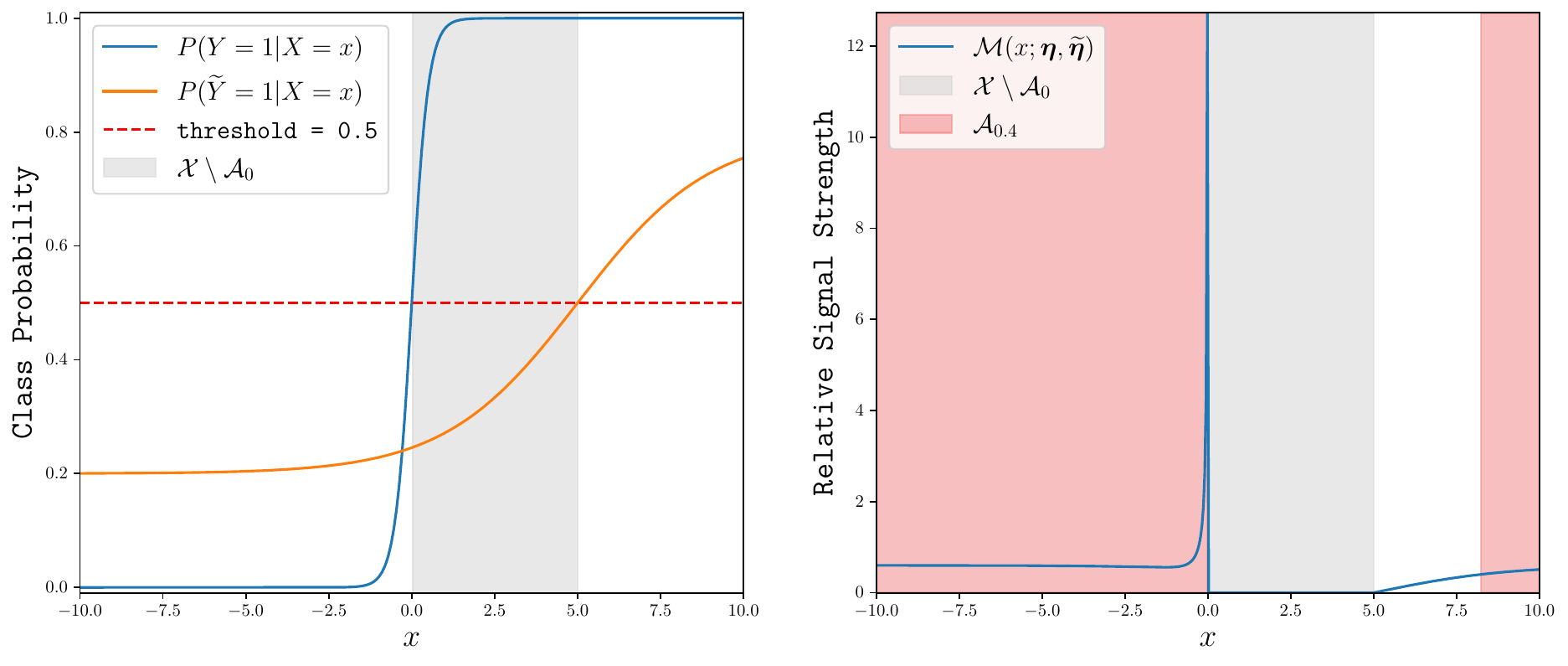}
    \caption{Illustration of relative signal strength for binary classification. \emph{Left}: clean and noisy posteriors $[\veta(x)]_1 = \prob{Y = 1 | X = x}$ and $[\widetilde{\veta}(x)]_1 = \bb{P} \big( \widetilde{Y} = 1 | X = x \big)$.
    \emph{Right}: relative signal strength corresponding to these posteriors. 
    The gray region, $x \in (0, 5)$, is where the true and noisy Bayes classifiers differ, and is also the zero signal region $\mc{X} \setminus \mc{A}_0$. The red region is $\mc{A}_{0.4}$, where the RSS is $> 0.4$. Note that as $x \uparrow 0, \mc{M}(x;\veta, \tveta) \uparrow \infty$, which occurs since $[\veta(x)]_1 \uparrow 1/2,$ while $[\widetilde{\veta}]_1$ is far from $1/2$. For $x = 0^+,$ the predicted labels under $\veta$ and $\widetilde{\veta}$ disagree, and the RSS crashes to $0.$
    }
    \label{fig:relative_signal_strength_example}
\end{figure}

\end{example}

\subsection{Posterior Drift Model Class.} 

Now putting definitions together, 
we consider the posterior drift model $\Pi$ defined over the triple $(P_X, \veta, \widetilde{\veta})$.
Let $\epsilon \in [0, 1] , \kappa \in [0, + \infty)$, and define
\begin{align*}
    \Pi(\epsilon, \kappa) 
    := \Big\{ 
    \left( P_X, \veta, \widetilde{\veta} \right): 
    P_X \Big( \mc{A}_{\kappa} \left(\veta, \widetilde{\veta} \right) \Big) \geq 1 - \epsilon
    \Big\}.
\end{align*}
This is a set of triples (label noise problems) such that $\mc{A}_{\kappa}$, the region with RSS at least $\kappa$, covers at least $1 - \epsilon$ of the probability mass. 
In the next section, we will demonstrate that the performance within $\mc{A}_{\kappa}$ can be guaranteed, whereas learning outside the region $\mc{A}_{\kappa}$ is provably challenging. 


\section{Upper and lower bounds}
\label{sec:bounds}

In this section, we establish  both upper and lower bounds for excess risk under multi-class instance-dependent label noise. 

\subsection{Minimax lower bound}

Our first theorem reveals a fundamental limit: no classifier trained using noisy data can surpass the constraints imposed by relative signal strength in a minimax sense. To state the theorem, we employ the following notation and terminology. Denote the noisy training data by $ Z^n = \big\{ ( X_i, \widetilde{Y}_i ) \big\}_{i=1}^n \stackrel{i.i.d.}{\sim} P_{X\widetilde{Y}}$. A \emph{learning rule} $\hat{f}$ is an algorithm that takes $Z^n$ and outputs a classifier. The risk $R(\hat{f})$ of a learning rule is a random variable, where the randomness is due to the draw $Z^n$.


\begin{theorem}[Minimax Lower Bound]
    Let $\epsilon \in [0, 1] , \kappa > 0$. 
    Then
    \begin{align*}
        \inf_{\hat{f}} 
        \sup_{ (P_X, \veta, \widetilde{\veta}) \in \Pi(\epsilon, \kappa) } \expect{ Z^n }{ R \left( \hat{f} \right)  - R(f^*) } \geq \frac{K-1}{K}  \epsilon + \Omega \left( \frac{1}{\kappa} \ \sqrt{\frac{1}{n}} \right),
    \end{align*}
    where the $\inf$ is over all learning rules.
    \label{thm:minimax_pd}
\end{theorem}



The proof in \Cref{subsec:minimax_proof} offers insights into how label noise impacts the learning process: On the low RSS region ($\mathcal{X}\backslash \mathcal{A}_\kappa$), learning is difficult if not impossible, and the learner incurs an irreducible error of $(1-1/K)\epsilon$. On the high RSS region ($\mathcal{A}_\kappa$), the standard nonparametric rate 
\citep{devroye1996probabilistic} is scaled by $1/\kappa$. These aspects determine fundamental limits that no classifier trained only on noisy data can overcome without additional assumptions.

\subsection{Upper bound}\label{sec:upper_bounds}

This subsection establishes an upper bound for
\emph{Noise Ignorant Empirical Risk Minimizer (NI-ERM)}, 
the empirical risk minimizer trained on noisy data. This result implies that NI-ERM is (nearly) minimax optimal, a potentially surprising result given that NI-ERM is arguably the simplest approach one might consider.
We begin by presenting a general result on the excess risk of any classifier, which is proved in Appendix \ref{subsec:oracle_proof}.

\begin{lemma}[Oracle Inequality]
    For any label noise problem $(P_X, \veta, \widetilde{\veta}) $ and any classifier $f$, 
    \begin{align*}
        R(f) - R(f^*) \leq \inf_{\kappa > 0} \left\{ P_X \Big( \mc{X} \setminus \mc{A}_{\kappa} \left(\veta, \widetilde{\veta} \right) \Big) + \frac{1}{\kappa} \left( \widetilde{R}(f) - \widetilde{R} \left( \tilde{f}^* \right) \right) \right\} .
    \end{align*}
    \label{lem:excess_risk}
\end{lemma}



For $(P_X, \veta, \widetilde{\veta}) \in \Pi(\epsilon, \kappa)$, the first term is bounded by $\epsilon$. When $f$ is selected by ERM over the noisy training data, conventional learning theory implies a bound on the second term. This leads to the following upper bound for NI-ERM, whose proof is in Appendix \ref{subsec:erm_generalization_proof}.

\begin{theorem}[Excess Risk Upper Bound of NI-ERM] 
    Let $\epsilon \in [0, 1] , \kappa > 0$.
    Consider any $(P_X, \veta, \widetilde{\veta}) \in \Pi(\epsilon, \kappa)$, assume function class $\mathcal{F}$ has Natarajan dimension $V$, and the noisy Bayes classifier $\tilde{f}^*$ belongs to $\mathcal{F}$.
    Let $\hat{f} \in \mathcal{F}$ be the ERM trained on $ Z^n = \big\{ ( X_i, \widetilde{Y}_i ) \big\}_{i=1}^n $, i.e., $\hat{f} = \argmin{f \in \mc{F}} \frac{1}{n} \sum_{i=1}^n \indicator{ \left\{ f(X_i) \neq \widetilde{Y}_i \right\} } $.
    Then 
    \begin{align*}
        \expect{ Z^n }{ R \left( \hat{f} \right)  - R(f^*) } 
        \leq \epsilon + \tilde{\mc{O}} \left( \frac{1}{\kappa} \sqrt{\frac{V}{n}} \right). 
    \end{align*}

    \label{thm:erm_generalization}

\end{theorem}





$\tilde{\mc{O}}$ denotes big-$\mc{O}$ notation ignoring logarithmic factors.
The Natarajan dimension is a multiclass analogue of the VC dimension. The upper bound in \Cref{thm:erm_generalization} aligns with the minimax lower bound (\Cref{thm:minimax_pd}) in both terms. 
For the irreducible error \(\epsilon\), there is a small gap of \(1/K\). This gap arises because, in the lower bound construction, the low signal region \(\mc{X} \setminus \mc{A}_\kappa\) is known to the learner, whereas knowledge of \(\mc{X} \setminus \mc{A}_\kappa\) is not provided to NI-ERM. If $\mc{A}_\kappa$ were known to the learner (an unrealistic assumption), then a mixed strategy that preforms NI-ERM on $\mc{A}_\kappa$ and randomly guesses on \(\mc{X} \setminus \mc{A}_\kappa\) would have an upper bound with first term of $(1-1/K)\epsilon$, exactly matching the lower bound. Regarding the second term, there is a universal constant and a logarithmic factor between the lower and upper bounds, which is a standard outcome in learning theory.

This result is surprising as it indicates that the simplest possible approach, which ignores the presence of noise, is nearly optimal. No learning rule could perform significantly better in this minimax sense.



\subsection{A smooth margin-condition on the relative signal strength}
\label{subsec:smooth_margin}

The previous sections have analyzed learning with label noise over the class $\Pi(\epsilon, \kappa) = \{ (P_X, \veta, \tilde{\veta}) : P_X(\mc{A}_{\kappa}) \geq 1 - \epsilon \}.$ Now, the set $\mc{X} \setminus \mc{A}_{\kappa}$ equals $ (\mc{X} \setminus  \mc{A}_0) \cup (\mc{A}_0 \setminus \mc{A}_\kappa)$. The first part of this decomposition is the region where the Bayes classifiers under the noisy and clean distributions differ, while the second is a region where these match, but the RSS is small. Naturally, while $P_X(\mc{X} \setminus  \mc{A}_0)$ must be incurred as an irreducible error, one may question why the class $\Pi$ also limits the mass of $\mc{A}_0\setminus \mc{A}_\kappa$. After all, with enough data, the optimal prediction in this region can be learned.

This issue would be resolved if there existed a $\kappa_0 > 0$ such that $P_X(\mc{A}_0) = P_X(\mc{A}_{\kappa_0}),$ i.e., if $P_X(\mc{A}_0\setminus \mc{A}_{\kappa_0}) = 0$. In fact, our lower bound from Theorem~\ref{thm:minimax_pd} uses precisely such a construction. An interesting point of comparison to this condition lies in Massart's hard-margin condition from standard supervised learning theory, which, for binary problems, demands that $P_X( |[\veta(x)]_1 -[\veta(x)]_2|< h) = 0$ for some $h > 0$, under which one obtains minimax excess risk bounds of $ O( \nicefrac{V}{nh})$ \citep{massart2006risk}. With this lens, we can view the condition $P_X( \mc{A}_0 \setminus \mc{A}_{\kappa_0}) = 0$ as a type of \emph{hard-margin condition on the relative signal strength $\mc{M}$}. This naturally motivates a smoothened version of this condition, inspired by Tsybakov's soft-margin condition  \citep{tsybakov2004optimal}.

\begin{definition}
    A triple $(P_X, \veta, \widetilde\veta)$ satisfies an $(\epsilon, \alpha, C_{\alpha})$-smooth \emph{relative signal margin condition} with $\epsilon \in [0, 1], \alpha > 0, C_{\alpha} > 0$ if 
    \begin{align*}
         \forall \kappa > 0, \ P_X(\mc{M}(x;\veta,\widetilde\veta) \leq \kappa ) \le C_\alpha \kappa^\alpha + \epsilon.
    \end{align*}
    
    Further, we define ${\Pi}'(\epsilon, \alpha, C_{\alpha})$ as the set of triples $(P_X, \veta, \widetilde\veta)$ that satisfy an $(\epsilon,\alpha, C_{\alpha})$-smooth relative signal margin condition.
\end{definition}

We show in Appendix~\ref{subsec:smooth_margin_erm_proof} that the techniques of Section \ref{sec:upper_bounds} yield the following result for $\Pi'$.

\begin{theorem}\label{thm:risk_bound_smooth_margin}
    Let $\epsilon \in [0, 1] , \alpha > 0, C_{\alpha} > 0$. 
    Consider any $(P_X, \veta, \widetilde{\veta}) \in {\Pi}'(\epsilon, \alpha, C_{\alpha})$, assume function class $\mathcal{F}$ has Natarajan dimension $V$, and the noisy Bayes classifier $\tilde{f}^*$ belongs to $\mathcal{F}$.
    Let $\hat{f} \in \mathcal{F}$ be the ERM trained on $ Z^n = \big\{ ( X_i, \widetilde{Y}_i ) \big\}_{i=1}^n $.
    Then 
    \begin{align*}
        \expect{Z^n}{R \left( \hat{f} \right) - R\left(f^*\right)}  
         \le \epsilon + \inf_{\kappa > 0} \left\{ C_\alpha \kappa^\alpha + \tilde{\mathcal{O}}\left( \frac{1}{\kappa} \sqrt{\frac{V}{n}} \right) \right\} 
        = \epsilon + \tilde{\mathcal{O}} \left( n^{-\alpha/(2 + 2\alpha)} \right).
    \end{align*}
\end{theorem}


Compared to Theorem~\ref{thm:erm_generalization}, we see that the rate of the second term is slightly slower, which is consistent with standard learning theory where Massart's hard margin assumption leads to faster rates than Tsybakov's. The advantage of the smooth relative margin is that the irreducible term in the above theorem is exactly $P_X(\mc{X} \setminus \mc{A}_0)$,
which has a clearer meaning as it measures the mismatch between clean and noisy Bayes classifiers.  Further, notice that the NI-ERM algorithm does not need information about $\alpha$, and thus the result is adaptive to both $\alpha$, and to the optimal $\kappa$ for each value of $\alpha$, as a consequence of the oracle inequality of Lemma \ref{lem:excess_risk}.



More broadly, Theorem~\ref{thm:risk_bound_smooth_margin} illustrates the flexibility of our conceptualization of label noise problems through RSS. The RSS $\mc{M}$ characterizes the irreducible error in label noise learning, similar to how the regression function $\veta$ characterizes excess risk in standard learning. Thus, standard theoretical frameworks can be adapted to the noisy label problem via the \emph{relative signal}.

\section{Conditions that ensure noise immunity}
\label{sec:immunity}


The minimax lower bound in the previous section revealed a negative outcome, indicating that no method can do well in the low signal region. 
Nevertheless, numerous empirical successes have been observed even under significant label noise.
This is not mere coincidence. In this section, we will illustrate that the high signal region $\mc{A}_{\kappa}$ can indeed cover the entire input space $\mc{X}$ even under massive label noise, albeit with the constraint $\mc{A}_{\kappa} \subseteq \mc{A}_0$ as stated in Proposition \ref{prop:signal_subset}. This not only explains past empirical successes, but also gives a rigorous condition on the consistency of NI-ERM.



This section will delve into the study of noise transition matrix $\E$ and establish precise conditions that lead to $\mc{A}_0 = \mc{X}$. These conditions are linear algebraic conditions on $\E$ that
ensure $\arg \max \widetilde{\veta}(x) = \arg \max \veta(x)$.
As a result, we can infer that in a $10$-class classification problem, even with up to $90\%$ of training labels being incorrect, the NI-ERM can still asymptotically achieve Bayes accuracy.
In the upcoming definition, we introduce the concept of noise immunity, wherein the optimal classifiers remain unaffected by label noise \citep{menon2015learning, scott2019generalized}.

\begin{definition}[Immunity]
    We say that a $K$-class classification problem $(P_X, \veta, \widetilde{\veta})$  is \emph{immune to label noise} if \(\forall x \in \mc{X}, \arg\max \tveta(x) = \arg\max \veta(x). \)
\end{definition}

Notice that due to Proposition \ref{prop:signal_subset}, if a problem is immune, then $\mc{A}_0 = \mc{X}$. 
We now provide necessary and sufficient conditions on noise transition matrix $\E$ that ensure noise immunity. We begin by considering distribution $P_{XY}$ with zero Bayes risk, that is, where $\veta$ is one-hot almost surely. A matrix is defined as diagonally dominant if, for each row, the diagonal element is the unique maximum.

\begin{theorem}[Immunity for Zero-error Distribution]
If $P_{XY}$ has Bayes risk of zero, then immunity holds if and only if for all $x$, the noise transition matrix $\E(x)$ is diagonally dominant.
    \label{thm:immunity_onehot}
\end{theorem}

\begin{remark*}
    For a zero-error distribution $P_{XY}$, even corrupted with instance-dependent label noise, achieving the Bayes risk is still feasible with a noise rate $\bb{P} \big( \widetilde{Y} \neq Y \big)$ up to  \smash{$ \frac{K-1}{K}$}. 
    This highlights that the task of classification itself is robust to label noise, specially when the clean $\veta$ is well-separated.
\end{remark*}

The above result relies on strong assumptions about the distribution $P_{XY}$. Now, we present a result that applies to any distribution, which, as a trade-off, turns out to impose more requirements on $\E$.

\begin{theorem}[Universal Immunity] 
    For any choice of $P_{XY}$, immunity holds
    \begin{align*}
        \quad \Longleftrightarrow \quad
        \exists \ e(x) > 0 \text{ s.t. } \forall x \in \mc{X}, \
        [\E(x)]_{i,j} = 
        \begin{cases}
            \frac{1}{K} + e(x) & i = j \\
            \frac{1}{K} - \frac{e(x)}{K-1} & i \neq j.
        \end{cases}
    \end{align*}
    \label{thm:symmetric_noise}
\end{theorem}


Previous works \citep{ghosh2017robust, menon2018learning, oyen2022robustness} have established that symmetric label noise is sufficient for immunity. Our contribution advances this understanding by demonstrating that such noise conditions are not only sufficient but also necessary. Specifically, under symmetric label noise, learning towards the Bayes classifier is feasible as long as the proportion of wrong labels does not exceed $\frac{K-1}{K}$. Furthermore, this transition is abrupt: when $\bb{P} \big( \widetilde{Y} \neq Y \big) < \frac{K-1}{K}$, $\mc{A}_0 = \mc{X}$, but when $\bb{P} \big( \widetilde{Y} \neq Y \big) \geq \frac{K-1}{K}$, $\mc{A}_0 = \emptyset$. Consequently, we expect to see a sudden drop in performance when noise rate passes the threshold.

The rationale behind the necessity of $\E(x)$ taking this specific form is that it redistributes the probability mass of $\veta$ in a ``uniform'' manner. This constraint arises because $\E(x)$ cannot favor any classes besides the true class. For instance, consider $\veta(x) = \left[\frac{1}{K} + \delta \ \frac{1}{K}  - \delta \ \frac{1}{K} \cdots \frac{1}{K} \right]^\top$ for some small $\delta > 0$, a ``non-uniform'' $\E(x)$ would alter the $\arg \max$.

\begin{figure}[t]
    \centering
    \begin{subfigure}{0.45\textwidth}
        \centering
        \resizebox{\linewidth}{!}{
            \begin{tikzpicture}
            \begin{axis}[
                title={Binary},
                xlabel={Noise Rate},
                ylabel={Testing Accuracy},
                legend style={font=\footnotesize},
                ymajorgrids=true,
                grid style=dashed,
                xmin=0-0.05, xmax=1+0.05,
                ymin=0-0.05, ymax=1+0.05,
                legend pos= south west,
                legend style={font=\footnotesize},
            ]
            
            \addplot+[blue, mark = *, only marks] plot coordinates {
                (0, 0.93)
                (0.05, 0.92825)
                (0.1, 0.92375)
                (0.15, 0.92075)
                (0.2, 0.91775)
                (0.25, 0.91925)
                (0.3, 0.9225)
                (0.35, 0.929)
                (0.4, 0.91825)
                (0.45, 0.92575)
                (0.5, 0.74675)
                (0.55, 0.25125)
                (0.6, 0.097)
                (0.65, 0.07725)
                (0.7, 0.07525)
                (0.75, 0.0725)
                (0.8, 0.07375)
                (0.85, 0.0715)
                (0.9, 0.07175)
                (0.95, 0.071)
                (1, 0.07)
            };
            \addlegendentry{Linear}
            
            \addplot+[orange, dashed, thin, no markers] coordinates {
                (0, 0.93125)
                (0.5, 0.93125)
                (0.5, 0.06875)
                (1, 0.06875)
            };
            \addlegendentry{Theory}
            
            \end{axis}
            \end{tikzpicture}
        }
        \caption{Logistic regression trained on 2D gaussian mixture data with different levels of symmetric noise}
    \end{subfigure}
    \hfill
    \begin{subfigure}{0.45\textwidth}
        \centering
        \resizebox{\linewidth}{!}{
            \begin{tikzpicture}
        \begin{axis}[
            title={10-class},
            xlabel={Noise Rate},
            ylabel={Testing Accuracy},
            legend pos=south west,
            legend style={font=\footnotesize},
            ymajorgrids=true,
            grid style=dashed,
            xmin=0-0.05, xmax=1+0.05,
            ymin=0-0.05, ymax=1+0.05,
        ]
        
        \addplot+[only marks, mark=*, mark options={fill=blue}] coordinates {
            (0, 0.9897)
            (0.05, 0.9889)
            (0.1, 0.9881)
            (0.15, 0.9846)
            (0.2, 0.9849)
            (0.25, 0.9816)
            (0.3, 0.9846)
            (0.35, 0.9807)
            (0.4, 0.9798)
            (0.45, 0.9757)
            (0.5, 0.9788)
            (0.55, 0.9784)
            (0.6, 0.9719)
            (0.65, 0.971)
            (0.7, 0.967)
            (0.75, 0.9502)
            (0.8, 0.89)
            (0.85, 0.8372)
            (0.9, 0.1158)
            (0.95, 0.0017)
            (1, 0.0003)
        };
        \addlegendentry{2-layer CNN}
        
        \addplot+[orange, dashed, thin, no markers] coordinates {
            (0, 1)
            (0.9, 1)
            (0.9, 0)
            (1, 0)
        };
        \addlegendentry{Theory}
        
        \end{axis}
        \end{tikzpicture}
        }
        \caption{2-layer CNN trained on MNIST with different levels of symmetric noise}
    \end{subfigure}
    
    \caption{Data simulation that verifies noise immunity. For binary, the turning point is at noise rate $\bb{P} \big(\widetilde{Y} \neq Y \big) = 0.5$. For $10$-class, the turning point is at $\bb{P} \big(\widetilde{Y} \neq Y \big) = 0.9$.
    }
    \label{fig:immunity}
\end{figure}
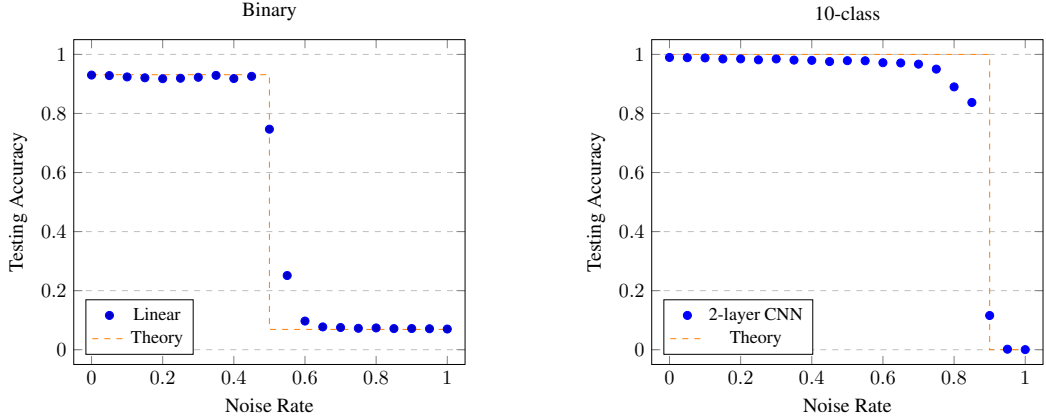

The above theorems demonstrate that signal strength at $x$ can still be high even under massive label noise $\bb{P} \big(\widetilde{Y} \neq Y\big)$, and, in essence, it is the discrete nature of the classification problem that allows robustness to label noise.
When immunity holds, the irreducible error in Theorem \ref{thm:erm_generalization} vanishes, therefore NI-ERM becomes a consistent learning rule. We validate this through data simulations presented in Figure \ref{fig:immunity}, where we systematically flip labels uniformly and observe the corresponding changes in the testing accuracy of NI-ERM. The simulation results align closely with the theoretical expectations: NI-ERM achieves near-Bayes risk performance until a certain noise threshold is reached, beyond which the testing performance sharply deteriorates.

    

\section{Practical implication 
}
\label{sec:practical}

The modern practice of machine learning often involves training a deep neural network. In complex tasks involving noisy labels, the na\"{i}ve NI-ERM is often outperformed by state-of-the-art methods by a significant extent \citep{li2020dividemix, xiao2023promix}. This is consistent with the finding that directly training a large neural network on noisy data frequently leads to overfitting \citep{zhang2021understanding}. 

Yet this is not grounds for abandoning NI-ERM altogether as a practical strategy. Instead of using NI-ERM for end-to-end training of a deep neural network, we instead propose the following simple, two-step procedure, termed `feature extraction + NI-ERM'.

\begin{enumerate}[leftmargin = 10pt, itemsep = 0.2\baselineskip, topsep =0.2\baselineskip, parsep = 0.2\baselineskip]
\item Perform feature extraction using any method (e.g., transfer learning or self-supervised learning) that does not require labels.
\item Learn a simple classifier (e.g., a linear classifier) on top of these extracted features, using the noisily labelled data, in a noise-ignorant way.
\end{enumerate}
This approach has three advantages over full network training. First, it avoids the potentially negative impact of the noisy labels on the extracted features. Second, it enjoys the inherent robustness of fitting a simple model (step 2) on noisy data, which we observed in Figure \ref{fig:immunity}. Third, it avoids the need to tune hyperparameters of the feature extractor using noisy labels. 
We note that a ``self-supervised + simple approach'' to learning was previously studied by \citet{bansal2021self}, although their focus was on generalisation properties without label noise. We also acknowledge that the practical idea of ignoring label noise is not new \citep{ghosh2021contrastive}, but the full power of this approach has not been previously recognized. 
For example, prior works often combine this approach with additional steps or employ early stopping to mitigate the effects of noise \citep{zheltonozhskii2022contrast, xue2022investigating}.

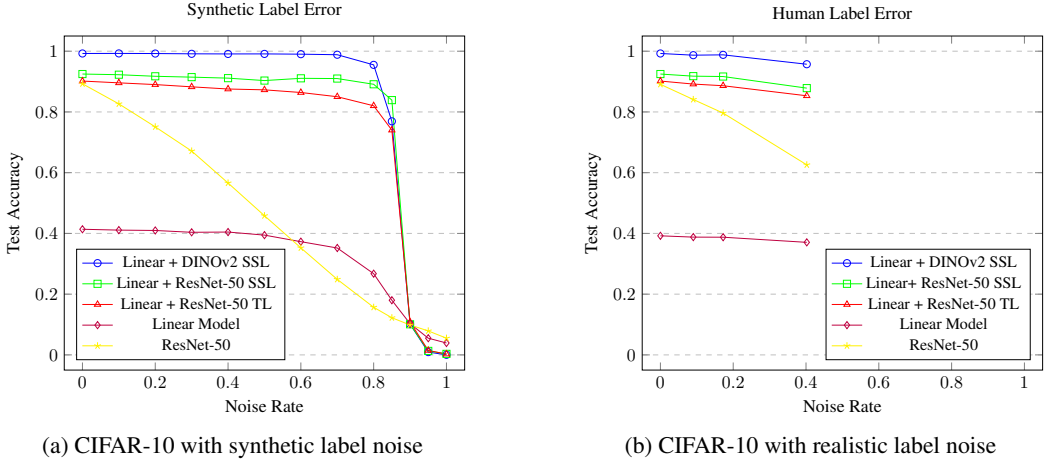
\begin{figure}[h]
    \centering
    \begin{subfigure}{0.45\textwidth}
        \centering
        \resizebox{\linewidth}{!}{
            \begin{tikzpicture}
                \begin{axis}[
                    title={Synthetic Label Error},
                    xlabel={Noise Rate},
                    ylabel={Test Accuracy},
                    xmin=-0.05, xmax=1+0.05,
                    ymax=1+0.05, ymin=0-0.05,
                    height=1.3\textwidth, 
                    legend pos= south west,
                    legend style={font=\footnotesize},
                    ymajorgrids=true,
                    grid style=dashed,
                ]


                \addplot[
                    color = blue,
                    mark = o,
                    ]
                    coordinates {
                    (0, 99.25/100) (0.1, 99.27/100)  (0.2, 99.23/100) (0.3, 99.14/100) (0.4, 99.1/100) (0.5, 99.11/100) (0.6, 99.02/100) (0.7, 98.84/100) (0.8, 95.5/100) (0.85, 76.91/100) (0.9, 10.13/100) (0.95, 0.92/100) (1.0, 0.03/100)
                    };
                    \addlegendentry{Linear + DINOv2 SSL}
            
                \addplot[
                    color = green,
                    mark = square,
                    ]
                    coordinates { 
                    (0, 92.48/100) (0.1, 92.26/100) (0.2, 91.74/100) 
                    (0.3, 91.46/100) (0.4, 91.13/100) (0.5, 90.33/100) 
                    (0.6, 91.07/100) (0.7, 90.99/100) (0.8, 89.11/100) 
                    (0.85, 83.89/100) (0.9, 10.08/100) (0.95, 1.31/100) 
                    (1.0, 0.34/100)
                    };
                    \addlegendentry{Linear + ResNet-50 SSL}

                \addplot[
                    color = red,
                    mark = triangle,
                    ]
                    coordinates {
                    (0, 90.17/100) (0.1, 89.58/100) (0.2, 89.01/100)
                    (0.3, 88.27/100) (0.4, 87.55/100) (0.5, 87.28/100)
                    (0.6, 86.4/100) (0.7, 85.01/100) (0.8, 82.03/100)
                    (0.85, 74.02/100) (0.9, 10.82/100) (0.95, 1.47/100)
                    (1.0, 0.26/100)
                    };
                    \addlegendentry{Linear + ResNet-50 TL}

                \addplot[
                    color = purple,
                    mark = diamond,
                    ]
                    coordinates {
                    (0, 41.37/100) (0.1, 41.09/100) (0.2, 40.97/100)
                    (0.3, 40.37/100) (0.4, 40.45/100) (0.5, 39.44/100)
                    (0.6, 37.28/100) (0.7, 35.2/100) (0.8, 26.74/100)
                    (0.85, 18/100) (0.9, 10.28/100) (0.95, 5.5/100)
                    (1.0, 3.92/100)
                    };
                    \addlegendentry{Linear Model}

                \addplot[
                    color = yellow,
                    mark = star,
                    ]
                    coordinates {
                    (0, 89.27/100) (0.1, 82.66/100) (0.2, 75.09/100) (0.3, 67.13/100) (0.4, 56.60/100) (0.5, 45.83/100) (0.6, 35.24/100) (0.7, 24.87/100) (0.8, 15.68/100) (0.85, 12.23/100) (0.9, 9.86/100)  (0.95, 7.89/100) (1.0, 5.50/100)
                    };
                    \addlegendentry{ResNet-50}

                \end{axis}
            \end{tikzpicture}
        }
        \caption{CIFAR-10 with synthetic label noise}
    \end{subfigure}
    \hfill
    \begin{subfigure}{0.45\textwidth}
        \centering
        \resizebox{\linewidth}{!}{
            \begin{tikzpicture}
                \begin{axis}[
                    title={Human Label Error},
                    xlabel={Noise Rate},
                    ylabel={Test Accuracy},
                    xmin=-0.05, xmax=1+0.05,
                    ymax=1+0.05, ymin=0-0.05,
                    height=1.3\textwidth, 
                    legend pos= south east,
                    legend style={font=\footnotesize},
                    ymajorgrids=true,
                    grid style=dashed,
                ]

                \addplot[
                    color = blue,
                    mark = o,
                    ]
                    coordinates {
                    (0, 99.25/100) (0.0903, 98.69/100) (0.1723, 98.8/100) (0.4021, 95.71/100)
                    };
                    \addlegendentry{Linear + DINOv2 SSL}
            
                \addplot[
                    color = green,
                    mark = square,
                    ]
                    coordinates {
                    (0, 92.48/100) (0.0903, 91.78/100) (0.1723, 91.66/100) (0.4021, 87.84/100)
                    };
                    \addlegendentry{Linear+ ResNet-50 SSL}

                \addplot[
                    color = red,
                    mark = triangle,
                    ]
                    coordinates {
                    (0, 90.17/100) (0.0903, 89.18/100) (0.1723, 88.63/100) (0.4021, 85.32/100)
                    };
                    \addlegendentry{Linear + ResNet-50 TL}

                \addplot[
                    color = purple,
                    mark = diamond,
                    ]
                    coordinates {
                    (0, 0.3921) (0.0903, 0.3878) (0.1723, 0.3875) (0.4021, 0.3706)
                    };
                    \addlegendentry{Linear Model}

                \addplot[
                    color=yellow,
                    mark=star,
                    ]
                    coordinates {
                    (0, 89.09/100) (0.0903, 84.10/100) (0.1723, 79.62/100) (0.4021, 62.58/100)
                    };
                    \addlegendentry{ResNet-50}

                \end{axis}
            \end{tikzpicture}
        }
        \caption{CIFAR-10 with realistic label noise}
    \end{subfigure}
    \caption{
    A linear model trained on features obtained from either transfer learning (pretrained ResNet-50 on ImageNet 
    \citep{he2016deep}
    ), self-supervised learning (ResNet-50 trained on CIFAR-10 images with contrastive loss \citep{chen2020simple}), or a pretrained self-supervised foundation model DINOv2 
    \citep{oquab2023dinov2} 
    significantly boosts the performance of the original linear model. In contrast, full training of a ResNet-50 leads to overfitting.}
    \label{fig:cifar_result}
\end{figure}

Remarkably, this two-step approach attains extremely strong performance. We conducted experiments \footnote{Code is available at: \url{https://github.com/allan-z/label_noise_ignorance}.} on the CIFAR image data under two scenarios: synthetic label flipping (symmetric noise) and realistic human label errors \citep{wei2022learning}, as shown in Figure \ref{fig:cifar_result}. We examine three different feature extractors: the DINOv2 foundation model \citep{oquab2023dinov2}, ResNet-50 features extracted from training on ImageNet \citep{he2016deep}, and self-supervised ResNet-50 using contrastive loss \citep{chen2020simple}. We also compared to a simple linear model trained on the raw pixel intensities, and a ResNet-50 trained end-to-end.
We observed that ResNet-50 exhibits degrading performance with increasing noise, consistent with previous findings \citep{zhang2021understanding, mallinar2022benign}. The linear model demonstrates robustness to noise, but suffers from significant approximation error.

Conversely, the FE+NI-ERM approach enjoys the best of both worlds. Regardless of how the feature extraction is carried out, the resulting models exhibit robustness to label noise, while the overall accuracy depends entirely on the quality of the extracted features. This is illustrated in Figure~\ref{fig:cifar_result}, where the flatness of the accuracy curves as noise increases indicates the robustness, and the intercept at zero label noise is a measure of the feature quality. Importantly, this property holds true even under realistic label noise of CIFAR-N \citep{wei2022learning}. In fact, we find that using the DINOv2 \citep{oquab2023dinov2} extracted features in our FE+NI-ERM approach yields state of the art results on the CIFAR-10N and CIFAR-100N benchmarks, across the noise levels, as shown in Table~\ref{tab:cifarn_leaderboard}.

We emphasize that the only hyperparameters of our model are the hyperparameters of the linear classifier, which are tuned automatically using standard cross-validation on the noisy labels. This contrasts to the implementation of many methods on the CIFAR-N leaderboard (\url{http://noisylabels.com/}) \footnote{If the link is inaccessible, see the May 23, 2024 archive captured by Wayback Machine: \url{https://web.archive.org/web/20240523101740/http://noisylabels.com/}.}, where the hyperparameters are hard-coded. Furthermore, our approach does not rely on data augmentation. 
Additional experiments, detailed in \Cref{subsec:additional_experiments}
, include comparisons with the `linear probing, then fine-tuning' approach \citep{kumar2022fine}, the application of different robust learning strategies on DINOv2 features, and results on synthetic instance-dependent label noise.  

Overall, the strong performance, the simplicity of the approach and the lack of any untunable hyperparameters highlights the effectiveness of FE+NI-ERM, and indicates the value of further investigation into its properties.

\begin{table}[htbp]
\centering
\caption{Performance comparison with CIFAR-N leaderboard (\url{http://noisylabels.com/}) in terms of testing accuracy. ``Aggre'', ``Rand1'', \dots, ``Noisy'' denote various types of human label noise. We compare with four methods that covers the top three performance for all noise categories: ProMix \citep{xiao2023promix}, ILL \citep{chen2023imprecise}, PLS \citep{albert2023your} and DivideMix \citep{li2020dividemix}.
Our approach, a Noise Ignorant linear model trained on features extracted by the self-supervised foundation model DINOv2 \citep{oquab2023dinov2}  achieves new state-of-the-art results, highlighted in bold. We employed Python's sklearn logistic regression and cross-validation functions without data augmentation; the results are deterministic and directly reproducible.}
\label{tab:cifarn_leaderboard}
\vspace{1em}

\resizebox{\textwidth}{!}{
\begin{tabular}{@{}ccccccc@{}}
\toprule
Leaderboard     & \multicolumn{5}{c}{CIFAR-10N}            & CIFAR-100N \\ \cmidrule(lr){2-6} \cmidrule(l){7-7}
Methods    & Aggre & Rand1 & Rand2 & Rand3 & Worst & Noisy \\ \midrule
 ProMix & 97.65 $\pm$ 0.19 & 97.39 $\pm$ 0.16 & 97.55 $\pm$ 0.12 & 97.52 $\pm$ 0.09 & \textbf{96.34 $\pm$ 0.23} & 73.79 $\pm$ 0.28  \\ 
 ILL  & 96.40 $\pm$ 0.03 & 96.06 $\pm$ 0.07 & 95.98 $\pm$ 0.12 & 96.10 $\pm$ 0.05 & 93.55 $\pm$ 0.14  & 68.07 $\pm$ 0.33   \\
 PLS  & 96.09 $\pm$ 0.09 & 95.86 $\pm$ 0.26 & 95.96 $\pm$ 0.16 & 96.10 $\pm$ 0.07 & 93.78 $\pm$ 0.30 & 73.25 $\pm$ 0.12   \\ 
 DivideMix & 95.01 $\pm$ 0.71 & 95.16 $\pm$ 0.19 & 95.23 $\pm$ 0.07 & 95.21 $\pm$ 0.14 & 92.56 $\pm$ 0.42 & 71.13 $\pm$ 0.48 \\
 \midrule
FE + NI-ERM & \textbf{98.69 $\pm$ 0.00} & \textbf{98.80 $\pm$ 0.00} & \textbf{98.65 $\pm$ 0.00} & \textbf{98.67 $\pm$ 0.00} & 95.71 $\pm$ 0.00 & \textbf{83.17 $\pm$ 0.00} \\ \bottomrule
\end{tabular}
}


\end{table}

\paragraph{RSS for realistic human label error.}

\begin{wrapfigure}{r}{0.55\linewidth}
    \centering
    \includegraphics[width=0.85\linewidth]{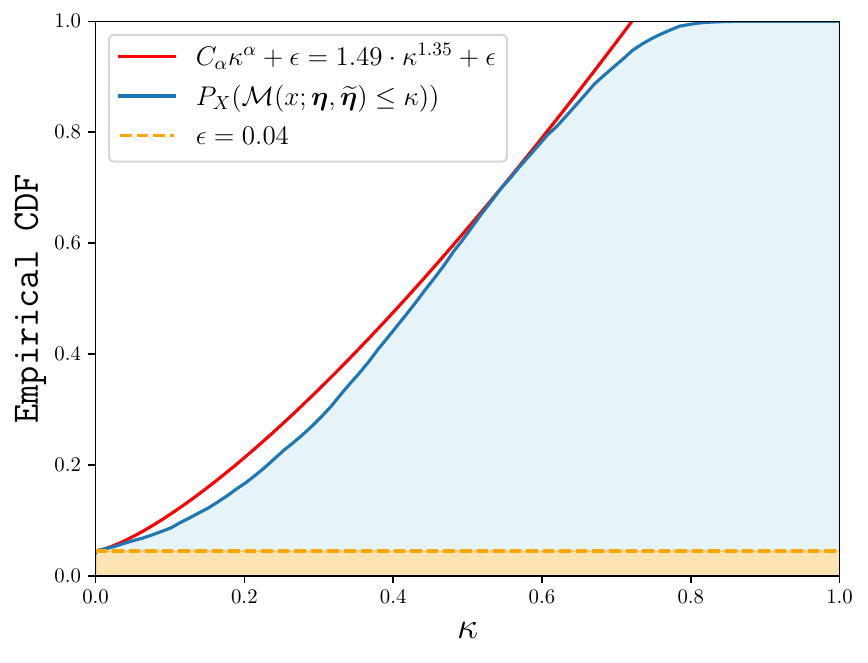}
    \caption{Empirical CDF of estimated RSS for CIFAR-10N, evaluated on test data.
    }
    \label{fig:realistic_rss}
\end{wrapfigure}

To calculate the RSS under realistic human label error, we train two linear classifiers on DINOv2 features under clean and noisy labels and use the models' predictions as estimates for the class probabilities $\veta$ and $\widetilde{\veta}$. Despite the high overall noise rate in CIFAR-10N ``Worst'' labels, with $\bb{P}(Y \neq \widetilde{Y}) = 40.21 \%$, we conjecture that the region where there is no signal, $\mc{X} \setminus \mc{A}_0$, covers only a small portion of the probability mass ($\epsilon \leq 4 \%$). Furthermore, the cumulative distribution of the estimated RSS can be upper-bounded by a polynomial $C_{\alpha} \kappa^{\alpha} + \epsilon$, supporting the validity of the smooth relative signal margin condition introduced in \Cref{subsec:smooth_margin}.

\section{Conclusions}
This work presents a rigorous theory for learning under multi-class, instance-dependent label noise. We establish nearly matching upper and lower bounds for excess risk and identify precise conditions for classifier consistency. 
Our theory reveals the (nearly) minimax optimality of Noise Ignorant Empirical Risk Minimizer (NI-ERM). To make this theory practical, we provide a simple modification leveraging a feature extractor with NI-ERM, demonstrating significant performance enhancements. A limitation of this work is that our methodology warrants more extensive experimental evaluation.

\newpage

\section*{Acknowledgements}

This work was supported in part by the National Science Foundation under award 2008074 and the Department of Defense, Defense Threat Reduction Agency under award HDTRA1-20-2-0002.
The authors thank Zixuan Huang, Yihao Xue for helpful discussions and Raj Rao Nadakuditi for feedback during a course project in which some early experiments in this paper were conducted. We also thank the anonymous reviewers for their suggestions, especially the reviewer who provided Example \ref{eg:KL_RSS}.

\bibliography{refs}

\begin{thebibliography}{52}
\providecommand{\natexlab}[1]{#1}
\providecommand{\url}[1]{\texttt{#1}}
\expandafter\ifx\csname urlstyle\endcsname\relax
  \providecommand{\doi}[1]{doi: #1}\else
  \providecommand{\doi}{doi: \begingroup \urlstyle{rm}\Url}\fi

\bibitem[Albert et~al.(2023)Albert, Arazo, Krishna, O’Connor, and McGuinness]{albert2023your}
Paul Albert, Eric Arazo, Tarun Krishna, Noel~E O’Connor, and Kevin McGuinness.
\newblock Is your noise correction noisy? pls: Robustness to label noise with two stage detection.
\newblock In \emph{Proceedings of the IEEE/CVF Winter Conference on Applications of Computer Vision}, pages 118--127, 2023.

\bibitem[Angluin and Laird(1988)]{angluin1988learning}
Dana Angluin and Philip Laird.
\newblock Learning from noisy examples.
\newblock \emph{Machine learning}, 2:\penalty0 343--370, 1988.

\bibitem[Bansal et~al.(2021)Bansal, Kaplun, and Barak]{bansal2021self}
Yamini Bansal, Gal Kaplun, and Boaz Barak.
\newblock For self-supervised learning, rationality implies generalization, provably.
\newblock In \emph{International Conference on Learning Representations}, 2021.

\bibitem[Brodley and Friedl(1999)]{brodley1999identifying}
Carla~E Brodley and Mark~A Friedl.
\newblock Identifying mislabeled training data.
\newblock \emph{Journal of artificial intelligence research}, 11:\penalty0 131--167, 1999.

\bibitem[Cai and Wei(2021)]{cai2021transfer}
T~Tony Cai and Hongji Wei.
\newblock Transfer learning for nonparametric classification: Minimax rate and adaptive classifier.
\newblock \emph{The Annals of Statistics}, 49\penalty0 (1):\penalty0 100--128, 2021.

\bibitem[Cannings et~al.(2020)Cannings, Fan, and Samworth]{cannings2020classification}
Timothy~I Cannings, Yingying Fan, and Richard~J Samworth.
\newblock Classification with imperfect training labels.
\newblock \emph{Biometrika}, 107\penalty0 (2):\penalty0 311--330, 2020.

\bibitem[Chen et~al.(2023)Chen, Shah, Wang, Tao, Wang, Xie, Sugiyama, Singh, and Raj]{chen2023imprecise}
Hao Chen, Ankit Shah, Jindong Wang, Ran Tao, Yidong Wang, Xing Xie, Masashi Sugiyama, Rita Singh, and Bhiksha Raj.
\newblock Imprecise label learning: A unified framework for learning with various imprecise label configurations.
\newblock \emph{arXiv preprint arXiv:2305.12715}, 2023.

\bibitem[Chen et~al.(2020)Chen, Kornblith, Norouzi, and Hinton]{chen2020simple}
Ting Chen, Simon Kornblith, Mohammad Norouzi, and Geoffrey Hinton.
\newblock A simple framework for contrastive learning of visual representations.
\newblock In \emph{International conference on machine learning}, pages 1597--1607. PMLR, 2020.

\bibitem[Cheng et~al.(2021)Cheng, Zhu, Li, Gong, Sun, and Liu]{cheng2021learning}
Hao Cheng, Zhaowei Zhu, Xingyu Li, Yifei Gong, Xing Sun, and Yang Liu.
\newblock Learning with instance-dependent label noise: A sample sieve approach.
\newblock In \emph{International Conference on Learning Representations}, 2021.

\bibitem[Deng et~al.(2009)Deng, Dong, Socher, Li, Li, and Fei-Fei]{deng2009imagenet}
Jia Deng, Wei Dong, Richard Socher, Li-Jia Li, Kai Li, and Li~Fei-Fei.
\newblock Imagenet: A large-scale hierarchical image database.
\newblock In \emph{2009 IEEE conference on computer vision and pattern recognition}, pages 248--255. IEEE, 2009.

\bibitem[Devroye et~al.(1996)Devroye, Gy{\"o}rfi, and Lugosi]{devroye1996probabilistic}
Luc Devroye, L{\'a}szl{\'o} Gy{\"o}rfi, and Gabor Lugosi.
\newblock \emph{A Probabilistic Theory of Pattern Recognition}, volume~31.
\newblock Springer, 1996.

\bibitem[Foret et~al.(2021)Foret, Kleiner, Mobahi, and Neyshabur]{foret2021sharpnessaware}
Pierre Foret, Ariel Kleiner, Hossein Mobahi, and Behnam Neyshabur.
\newblock Sharpness-aware minimization for efficiently improving generalization.
\newblock In \emph{International Conference on Learning Representations}, 2021.
\newblock URL \url{https://openreview.net/forum?id=6Tm1mposlrM}.

\bibitem[Ghosh and Kumar(2017)]{ghosh2017robust}
Aritra Ghosh and Himanshu Kumar.
\newblock Robust loss functions under label noise for deep neural networks.
\newblock In \emph{Proceedings of the AAAI conference on artificial intelligence}, 2017.

\bibitem[Ghosh and Lan(2021)]{ghosh2021contrastive}
Aritra Ghosh and Andrew Lan.
\newblock Contrastive learning improves model robustness under label noise.
\newblock In \emph{Proceedings of the IEEE/CVF Conference on Computer Vision and Pattern Recognition}, pages 2703--2708, 2021.

\bibitem[Ghosh et~al.(2015)Ghosh, Manwani, and Sastry]{ghosh2015making}
Aritra Ghosh, Naresh Manwani, and PS~Sastry.
\newblock Making risk minimization tolerant to label noise.
\newblock \emph{Neurocomputing}, 160:\penalty0 93--107, 2015.

\bibitem[Han et~al.(2018)Han, Yao, Yu, Niu, Xu, Hu, Tsang, and Sugiyama]{han2018coteaching}
Bo~Han, Quanming Yao, Xingrui Yu, Gang Niu, Miao Xu, Weihua Hu, Ivor Tsang, and Masashi Sugiyama.
\newblock Co-teaching: Robust training of deep neural networks with extremely noisy labels.
\newblock \emph{Advances in neural information processing systems}, 31, 2018.

\bibitem[He et~al.(2016)He, Zhang, Ren, and Sun]{he2016deep}
Kaiming He, Xiangyu Zhang, Shaoqing Ren, and Jian Sun.
\newblock Deep residual learning for image recognition.
\newblock In \emph{Proceedings of the IEEE conference on computer vision and pattern recognition}, pages 770--778, 2016.

\bibitem[Im and Grigas(2023)]{im2023binary}
Hyungki Im and Paul Grigas.
\newblock Binary classification with instance and label dependent label noise.
\newblock \emph{arXiv preprint arXiv:2306.03402}, 2023.

\bibitem[Kumar et~al.(2022)Kumar, Raghunathan, Jones, Ma, and Liang]{kumar2022fine}
Ananya Kumar, Aditi Raghunathan, Robbie Jones, Tengyu Ma, and Percy Liang.
\newblock Fine-tuning can distort pretrained features and underperform out-of-distribution.
\newblock In \emph{International Conference on Learning Representations}, 2022.

\bibitem[Li et~al.(2020)Li, Socher, and Hoi]{li2020dividemix}
Junnan Li, Richard Socher, and Steven~CH Hoi.
\newblock Dividemix: Learning with noisy labels as semi-supervised learning.
\newblock In \emph{International Conference on Learning Representations}, 2020.

\bibitem[Liu et~al.(2020)Liu, Niles-Weed, Razavian, and Fernandez-Granda]{liu2020early}
Sheng Liu, Jonathan Niles-Weed, Narges Razavian, and Carlos Fernandez-Granda.
\newblock Early-learning regularization prevents memorization of noisy labels.
\newblock \emph{Advances in neural information processing systems}, 33:\penalty0 20331--20342, 2020.

\bibitem[Liu et~al.(2022)Liu, Zhu, Qu, and You]{liu2022robust}
Sheng Liu, Zhihui Zhu, Qing Qu, and Chong You.
\newblock Robust training under label noise by over-parameterization.
\newblock In \emph{International Conference on Machine Learning}, pages 14153--14172. PMLR, 2022.

\bibitem[Liu and Tao(2015)]{liu2015classification}
Tongliang Liu and Dacheng Tao.
\newblock Classification with noisy labels by importance reweighting.
\newblock \emph{IEEE Transactions on pattern analysis and machine intelligence}, 38\penalty0 (3):\penalty0 447--461, 2015.

\bibitem[Liu and Guo(2020)]{liu2020peer}
Yang Liu and Hongyi Guo.
\newblock Peer loss functions: Learning from noisy labels without knowing noise rates.
\newblock In \emph{International conference on machine learning}, pages 6226--6236. PMLR, 2020.

\bibitem[Ma and Fattahi(2022)]{ma2022blessing}
Jianhao Ma and Salar Fattahi.
\newblock Blessing of depth in linear regression: Deeper models have flatter landscape around the true solution.
\newblock \emph{Advances in Neural Information Processing Systems}, 35:\penalty0 34334--34346, 2022.

\bibitem[Maity et~al.(2023)Maity, Dutta, Terhorst, Sun, and Banerjee]{maity2021linear}
Subha Maity, Diptavo Dutta, Jonathan Terhorst, Yuekai Sun, and Moulinath Banerjee.
\newblock {A linear adjustment based approach to posterior drift in transfer learning}.
\newblock \emph{Biometrika}, 2023.
\newblock URL \url{https://doi.org/10.1093/biomet/asad029}.

\bibitem[Mallinar et~al.(2022)Mallinar, Simon, Abedsoltan, Pandit, Belkin, and Nakkiran]{mallinar2022benign}
Neil Mallinar, James Simon, Amirhesam Abedsoltan, Parthe Pandit, Misha Belkin, and Preetum Nakkiran.
\newblock Benign, tempered, or catastrophic: Toward a refined taxonomy of overfitting.
\newblock \emph{Advances in Neural Information Processing Systems}, 35:\penalty0 1182--1195, 2022.

\bibitem[Massart and N{\'e}d{\'e}lec(2006)]{massart2006risk}
Pascal Massart and {\'E}lodie N{\'e}d{\'e}lec.
\newblock Risk bounds for statistical learning.
\newblock \emph{The Annals of Statistics}, 34\penalty0 (5):\penalty0 2326--2366, 2006.

\bibitem[Menon et~al.(2015)Menon, Van~Rooyen, Ong, and Williamson]{menon2015learning}
Aditya Menon, Brendan Van~Rooyen, Cheng~Soon Ong, and Bob Williamson.
\newblock Learning from corrupted binary labels via class-probability estimation.
\newblock In \emph{International conference on machine learning}, pages 125--134. PMLR, 2015.

\bibitem[Menon et~al.(2018)Menon, Van~Rooyen, and Natarajan]{menon2018learning}
Aditya~Krishna Menon, Brendan Van~Rooyen, and Nagarajan Natarajan.
\newblock Learning from binary labels with instance-dependent noise.
\newblock \emph{Machine Learning}, 107:\penalty0 1561--1595, 2018.

\bibitem[Natarajan(1989)]{natarajan1989learning}
Balas~K Natarajan.
\newblock On learning sets and functions.
\newblock \emph{Machine Learning}, 4:\penalty0 67--97, 1989.

\bibitem[Natarajan et~al.(2013)Natarajan, Dhillon, Ravikumar, and Tewari]{natarajan2013learning}
Nagarajan Natarajan, Inderjit~S Dhillon, Pradeep~K Ravikumar, and Ambuj Tewari.
\newblock Learning with noisy labels.
\newblock \emph{Advances in neural information processing systems}, 26, 2013.

\bibitem[Northcutt et~al.(2021)Northcutt, Jiang, and Chuang]{northcutt2021confident}
Curtis Northcutt, Lu~Jiang, and Isaac Chuang.
\newblock Confident learning: Estimating uncertainty in dataset labels.
\newblock \emph{Journal of Artificial Intelligence Research}, 70:\penalty0 1373--1411, 2021.

\bibitem[Oquab et~al.(2023)Oquab, Darcet, Moutakanni, Vo, Szafraniec, Khalidov, Fernandez, HAZIZA, Massa, El-Nouby, et~al.]{oquab2023dinov2}
Maxime Oquab, Timoth{\'e}e Darcet, Th{\'e}o Moutakanni, Huy~V Vo, Marc Szafraniec, Vasil Khalidov, Pierre Fernandez, Daniel HAZIZA, Francisco Massa, Alaaeldin El-Nouby, et~al.
\newblock Dinov2: Learning robust visual features without supervision.
\newblock \emph{Transactions on Machine Learning Research}, 2023.

\bibitem[Oyen et~al.(2022)Oyen, Kucer, Hengartner, and Singh]{oyen2022robustness}
Diane Oyen, Michal Kucer, Nicolas Hengartner, and Har~Simrat Singh.
\newblock Robustness to label noise depends on the shape of the noise distribution.
\newblock \emph{Advances in Neural Information Processing Systems}, 35:\penalty0 35645--35656, 2022.

\bibitem[Patrini et~al.(2017)Patrini, Rozza, Krishna~Menon, Nock, and Qu]{patrini2017making}
Giorgio Patrini, Alessandro Rozza, Aditya Krishna~Menon, Richard Nock, and Lizhen Qu.
\newblock Making deep neural networks robust to label noise: A loss correction approach.
\newblock In \emph{Proceedings of the IEEE conference on computer vision and pattern recognition}, pages 1944--1952, 2017.

\bibitem[Scott(2019)]{scott2019generalized}
Clayton Scott.
\newblock A generalized {N}eyman-{P}earson criterion for optimal domain adaptation.
\newblock In \emph{Algorithmic Learning Theory}, pages 738--761. PMLR, 2019.

\bibitem[Scott et~al.(2013)Scott, Blanchard, and Handy]{scott2013classification}
Clayton Scott, Gilles Blanchard, and Gregory Handy.
\newblock Classification with asymmetric label noise: Consistency and maximal denoising.
\newblock In \emph{Conference on learning theory}, pages 489--511. PMLR, 2013.

\bibitem[Tsybakov(2004)]{tsybakov2004optimal}
Alexander~B Tsybakov.
\newblock Optimal aggregation of classifiers in statistical learning.
\newblock \emph{The Annals of Statistics}, 32\penalty0 (1):\penalty0 135--166, 2004.

\bibitem[Van~Rooyen and Williamson(2018)]{van2018theory}
Brendan Van~Rooyen and Robert~C Williamson.
\newblock A theory of learning with corrupted labels.
\newblock \emph{Journal of Machine Learning Research}, 18\penalty0 (228):\penalty0 1--50, 2018.

\bibitem[Vapnik and Chervonenkis(1971)]{vapnik1971}
V.~N. Vapnik and A.~Ya. Chervonenkis.
\newblock On the uniform convergence of relative frequencies of events to their probabilities.
\newblock \emph{Theory of Probability \& Its Applications}, 16\penalty0 (2):\penalty0 264--280, 1971.
\newblock \doi{10.1137/1116025}.
\newblock URL \url{https://doi.org/10.1137/1116025}.

\bibitem[Wang et~al.(2022)Wang, Wang, and Liu]{wang2022estimating}
Jialu Wang, Eric~Xin Wang, and Yang Liu.
\newblock Estimating instance-dependent label-noise transition matrix using a deep neural network.
\newblock In \emph{International Conference on Machine Learning}, 2022.

\bibitem[Wei et~al.(2022)Wei, Zhu, Cheng, Liu, Niu, and Liu]{wei2022learning}
Jiaheng Wei, Zhaowei Zhu, Hao Cheng, Tongliang Liu, Gang Niu, and Yang Liu.
\newblock Learning with noisy labels revisited: A study using real-world human annotations.
\newblock In \emph{International Conference on Learning Representations}, 2022.

\bibitem[Xia et~al.(2020)Xia, Liu, Han, Wang, Gong, Liu, Niu, Tao, and Sugiyama]{xia2020part}
Xiaobo Xia, Tongliang Liu, Bo~Han, Nannan Wang, Mingming Gong, Haifeng Liu, Gang Niu, Dacheng Tao, and Masashi Sugiyama.
\newblock Part-dependent label noise: Towards instance-dependent label noise.
\newblock \emph{Advances in Neural Information Processing Systems}, 33:\penalty0 7597--7610, 2020.

\bibitem[Xiao et~al.(2023)Xiao, Dong, Wang, Feng, Wu, Chen, and Zhao]{xiao2023promix}
Ruixuan Xiao, Yiwen Dong, Haobo Wang, Lei Feng, Runze Wu, Gang Chen, and Junbo Zhao.
\newblock Promix: Combating label noise via maximizing clean sample utility.
\newblock In Edith Elkind, editor, \emph{Proceedings of the Thirty-Second International Joint Conference on Artificial Intelligence, {IJCAI-23}}, pages 4442--4450. International Joint Conferences on Artificial Intelligence Organization, 8 2023.
\newblock \doi{10.24963/ijcai.2023/494}.
\newblock URL \url{https://doi.org/10.24963/ijcai.2023/494}.
\newblock Main Track.

\bibitem[Xue et~al.(2022)Xue, Whitecross, and Mirzasoleiman]{xue2022investigating}
Yihao Xue, Kyle Whitecross, and Baharan Mirzasoleiman.
\newblock Investigating why contrastive learning benefits robustness against label noise.
\newblock In \emph{International Conference on Machine Learning}, pages 24851--24871. PMLR, 2022.

\bibitem[Yang et~al.(2023)Yang, Wu, Yang, Han, Liu, Xu, Niu, and Liu]{yang2023parametrical}
Shuo Yang, Songhua Wu, Erkun Yang, Bo~Han, Yang Liu, Min Xu, Gang Niu, and Tongliang Liu.
\newblock A parametrical model for instance-dependent label noise.
\newblock \emph{IEEE Transactions on Pattern Analysis and Machine Intelligence}, 2023.

\bibitem[Zhang et~al.(2021{\natexlab{a}})Zhang, Bengio, Hardt, Recht, and Vinyals]{zhang2021understanding}
Chiyuan Zhang, Samy Bengio, Moritz Hardt, Benjamin Recht, and Oriol Vinyals.
\newblock Understanding deep learning (still) requires rethinking generalization.
\newblock \emph{Communications of the ACM}, 64\penalty0 (3):\penalty0 107--115, 2021{\natexlab{a}}.

\bibitem[Zhang et~al.(2022)Zhang, Wang, and Scott]{zhang2022learning}
Jianxin Zhang, Yutong Wang, and Clayton Scott.
\newblock Learning from label proportions by learning with label noise.
\newblock \emph{Advances in Neural Information Processing Systems}, 35:\penalty0 26933--26942, 2022.

\bibitem[Zhang et~al.(2021{\natexlab{b}})Zhang, Lee, and Agarwal]{zhang2021learning}
Mingyuan Zhang, Jane Lee, and Shivani Agarwal.
\newblock Learning from noisy labels with no change to the training process.
\newblock In \emph{International Conference on Machine Learning}, pages 12468--12478. PMLR, 2021{\natexlab{b}}.

\bibitem[Zheltonozhskii et~al.(2022)Zheltonozhskii, Baskin, Mendelson, Bronstein, and Litany]{zheltonozhskii2022contrast}
Evgenii Zheltonozhskii, Chaim Baskin, Avi Mendelson, Alex~M Bronstein, and Or~Litany.
\newblock Contrast to divide: Self-supervised pre-training for learning with noisy labels.
\newblock In \emph{Proceedings of the IEEE/CVF Winter Conference on Applications of Computer Vision}, pages 1657--1667, 2022.

\bibitem[Zhu et~al.(2021)Zhu, Liu, and Liu]{zhu2021second}
Zhaowei Zhu, Tongliang Liu, and Yang Liu.
\newblock A second-order approach to learning with instance-dependent label noise.
\newblock In \emph{Proceedings of the IEEE/CVF conference on computer vision and pattern recognition}, pages 10113--10123, 2021.

\end{thebibliography}
\bibliographystyle{plainnat}

\newpage
\appendix

\section{Appendix / supplemental material}

\subsection{Equivalence of noise transition and domain adaptation perspectives}
\label{sec:equivalentperspective}

The noise transition perspective models the joint distribution of \((X, Y, \widetilde{Y})\), which can be characterized as:
\begin{align*}
    P_{X, Y, \widetilde{Y}} = P_X \underbrace{P_{Y|X}}_{\veta} \underbrace{P_{\widetilde{Y}|Y,X}}_{\E}
\end{align*}
Thus, by specifying \(P_X\), \(\veta\), and \(\E\), we obtain a triple \((P_X, \veta, \widetilde{\veta})\) with \(\widetilde{\veta}(x) = \E(x)^\top \veta(x)\).

In contrast, the domain adaptation perspective views label noise problems directly as a triple \((P_X, \veta, \widetilde{\veta})\), bypassing the explicit modeling of the noise transition matrix \(\E\). 

If no assumptions are made about the form of \(\E\), the domain adaptation view remains fully expressive. Given a triple \((P_X, \veta, \widetilde{\veta})\), we can always define a noise transition matrix as:
\begin{align*}
    \E(x) = \1 \widetilde{\veta}^\top,
\end{align*}
where \(\1 = [1 \dots 1]^\top\). We can verify that \(\E\) is row-stochastic, and
\begin{align*}
    \widetilde{\veta} &= \E(x)^\top \veta = (\widetilde{\veta} \1^\top) \veta = \widetilde{\veta} (\1^\top \veta) = \widetilde{\veta}.
\end{align*}
Therefore, these two perspectives are equivalent.

\subsection{Proofs}

\subsubsection{Proof of  \Cref{prop:signal_subset}}
\label{subsec:signal_subset_proof}

\begin{proposition*}
    \begin{align*}
        \mc{A}_{0} (\veta, \widetilde{\veta}) = \big\{x \in \mc{X}: \argmax{} \widetilde{\veta}(x) \subseteq \argmax{} {\veta}(x) \big\}. 
    \end{align*}
\end{proposition*}

\begin{proof}
    Notice that 
        \begin{align*}
        \mc{M}(x; \veta, \widetilde{\veta}) = 0 \quad \Longleftrightarrow{} \quad \argmax{} \widetilde{\veta}(x) \not\subseteq \argmax{} {\veta}(x).
    \end{align*}
    This is because $\mc{M}(x; \veta, \widetilde{\veta}) = 0$ when the numerator is zero and the denominator is non-zero, which happens when $\argmax{} \widetilde{\veta}(x) \not\subseteq \argmax{} {\veta}(x)$.
    
    An equivalent statement of this is
    \begin{align*}
        \mc{M}(x; \veta, \widetilde{\veta}) > 0 \quad \Longleftrightarrow{} \quad \argmax{} \widetilde{\veta}(x) \subseteq \argmax{} {\veta}(x).
    \end{align*}

\end{proof}

\subsubsection{Proof of \Cref{prop:rss_binary,}}

\begin{proposition*}
    In the binary setting, for $\kappa \geq 0$,
    \begin{align*}
        \mc{M}(x; \eta, \widetilde{\eta}) = \max{ \left\{ \frac{  \widetilde{\eta}(x) - \half }{ \eta(x) - \half } , 0 \right\} }, \qquad \text{and} \qquad
        \mc{A}_{\kappa} (\eta, \widetilde{\eta})  = \left\{ x \in \mc{X}: \frac{  \widetilde{\eta}(x) - \half }{ \eta(x) - \half } > \kappa \right\}.
    \end{align*}
\end{proposition*}

\begin{proof}
    In a brute-force way, we can examine the nine cases where $\eta(x), \widetilde{\eta}(x)$ is greater, equal, or smaller than $1/2$.

    If $\widetilde{\eta}(x) > \half, \eta(x) > \half$, then
    \begin{align*}
        \mc{M}(x; \veta, \widetilde{\veta}) 
        & = \min_{j \in \mc{Y}} \ \ \frac{ \max_i [\widetilde{\veta}(x)]_i - [\widetilde{\veta}(x)]_{j}  }{ \max_i [\veta(x)]_i - [\veta(x)]_{j} } \\
        & = \frac{ \widetilde{\eta}(x) - (1 - \widetilde{\eta}(x)) }{ \eta(x) - (1-\eta(x)) } \\
        & = \frac{  \widetilde{\eta}(x) - \half }{ \eta(x) - \half } \\
        & = \max{ \left\{ \frac{  \widetilde{\eta}(x) - \half }{ \eta(x) - \half } , 0 \right\} }.
    \end{align*}
    If $\widetilde{\eta}(x) < \half, \eta(x) < \half$, the same argument holds.

    If $\widetilde{\eta}(x) > \half, \eta(x) < \half$ or $\widetilde{\eta}(x) < \half, \eta(x) > \half$, take $j = \arg \max \widetilde{\veta}(x)$, we have
    \begin{align*}
        \mc{M}(x; \veta, \widetilde{\veta}) 
        & = \min_{j \in \mc{Y}} \ \ \frac{ \max_i [\widetilde{\veta}(x)]_i - [\widetilde{\veta}(x)]_{j}  }{ \max_i [\veta(x)]_i - [\veta(x)]_{j} } \\
        & = 0 \\
        & = \max{ \left\{ \frac{  \widetilde{\eta}(x) - \half }{ \eta(x) - \half } , 0 \right\} }.
    \end{align*}

    If $\widetilde{\eta}(x) = \half, \eta(x) < \half$ or $\widetilde{\eta}(x) = \half, \eta(x) < \half$, take $j \neq \arg \max \veta(x)$, we have
    \begin{align*}
        \mc{M}(x; \veta, \widetilde{\veta}) 
        & = \min_{j \in \mc{Y}} \ \ \frac{ \max_i [\widetilde{\veta}(x)]_i - [\widetilde{\veta}(x)]_{j}  }{ \max_i [\veta(x)]_i - [\veta(x)]_{j} } \\
        & = 0 \\
        & = \max{ \left\{ \frac{  \widetilde{\eta}(x) - \half }{ \eta(x) - \half } , 0 \right\} }.
    \end{align*}    
    If $\eta(x) = \half$, then 
    \begin{align*}
        \mc{M}(x; \veta, \widetilde{\veta}) 
        & = \min_{j \in \mc{Y}} \ \ \frac{ \max_i [\widetilde{\veta}(x)]_i - [\widetilde{\veta}(x)]_{j}  }{ \max_i [\veta(x)]_i - [\veta(x)]_{j} } \\
        & = \frac{\max_i [\widetilde{\veta}(x)]_i - [\widetilde{\veta}(x)]_{j}}{0} 
        & \forall j \\
        & = + \infty \\
        & = \max{ \left\{ \frac{  \widetilde{\eta}(x) - \half }{ \eta(x) - \half } , 0 \right\} }.
    \end{align*}  
    Note that it makes sense for RSS to be $+ \infty$ when $\eta(x) = \half$, because in this case, clean excess risk $R(f) - R(f^*)$ is $0$ at point $x$ for any classifier.

    Therefore, we can conclude that 
        \begin{align*}
        \mc{M}(x; \eta, \widetilde{\eta}) = \max{ \left\{ \frac{  \widetilde{\eta}(x) - \half }{ \eta(x) - \half } , 0 \right\} }, 
    \end{align*}
    by definition, we have, for $\kappa \geq 0$
    \begin{align*}
        \mc{A}_{\kappa} (\eta, \widetilde{\eta}) 
        & = \left\{ x \in \mc{X}: \mc{M}(x; \eta, \widetilde{\eta}) > \kappa \right\} \\
        & = \left\{ x \in \mc{X}: \frac{  \widetilde{\eta}(x) - \half }{ \eta(x) - \half } > \kappa \right\}.
    \end{align*}

    
\end{proof}

\subsubsection{Proof of lower bound: Theorem \ref{thm:minimax_pd}}
\label{subsec:minimax_proof}

Now we provide a more formal statement of the minimax lower bound and its proof. We begin with the scenario where the noisy distribution $P_{X \widetilde{Y}}$ has zero Bayes risk as an introductory example. The proof for the general case follows a similar strategy but involves more complex bounding techniques. We recommend that interested readers first review the proof of the zero-error version to build a solid understanding before tackling the general case.

Now consider a more restricted subset of $\Pi(\epsilon, \kappa)$:
\begin{align*}
    \Pi(\epsilon, \kappa, V, 0) 
    := \Big\{ 
    \left( P_X, \veta, \widetilde{\veta} \right): 
    P_X \Big( \mc{A}_{\kappa} \left(\veta, \widetilde{\veta} \right) \Big) \geq 1 - \epsilon, P_X \text{ supported on $V+1$ points}, \widetilde{R}^* = 0
    \Big\}.
\end{align*}

\begin{theorem*}[Minimax Lower Bound: when $\widetilde{R}^* = 0$]
    Let $\epsilon \in [0, 1] , \kappa > 0, V >1$. 
    For any learning rule $\hat{f}$ based upon $ Z^n = \big\{ ( X_i, \widetilde{Y}_i ) \big\}_{i=1}^n $, and $n > \max (V - 1, 2)$,

    \begin{align*}
        \sup_{ (P_X, \veta, \widetilde{\veta}) \in \Pi(\epsilon, \kappa)} \expect{ Z^n }{ R \left( \hat{f} \right)  - R(f^*) } 
        & \geq \sup_{ (P_X, \veta, \widetilde{\veta}) \in \Pi(\epsilon, \kappa, V, 0)} \expect{ Z^n }{ R \left( \hat{f} \right)  - R(f^*) } \\
        & \geq
        \frac{K-1}{K}  \epsilon + \frac{1}{\kappa}  \frac{(V-1)(1-\epsilon)}{8en} 
    \end{align*}
\end{theorem*}

\begin{proof}

    We will construct a triple $(P_X, \veta, \widetilde{\veta})$ that is parameterized by $j, \vb := [b_1 \ b_2 \ \cdots \ b_{V-1}]^\top$, and $\delta$.

    First, we define $P_X$. Pick any $V+1$ distinct points $x_0, x_1, \ldots, x_V$, 
    \begin{align*}
        P_X(x) = \begin{cases}
            \epsilon & x = x_0 \\
            (1 - \epsilon) \cdot \frac{1}{n} & x = x_1, \dots, x_{V-1} \\
            (1 - \epsilon) \cdot \left(1 - \frac{V-1}{n} \right) & x = x_V.
        \end{cases},
    \end{align*}
    this is where we need the condition that $n > V-1$.
    
    Then, define the clean and noisy class posteriors:    
    \begin{align}
        \text{If } x = x_0, & \text{ then } \  \veta(x) = \ve_j, \ \widetilde{\veta}(x) = \ve_1, \quad j \in \{1, 2, \dots K \} \label{eqn:minimax_zero_first_region} \\
        \text{If } x = x_t, &  \ 1 \leq t \leq V-1, \text{then} \
        \veta(x) = \begin{bmatrix}
            \frac{1}{2} + \frac{1}{2(\kappa + \delta)} \cdot (-1)^{b_t + 1} \\
            \frac{1}{2} - \frac{1}{2(\kappa + \delta)} \cdot (-1)^{b_t + 1} \\
            0 \\
            \vdots \\
            0
        \end{bmatrix}, \widetilde{\veta}(x) = \ve_{b_t}, b_t \in \{1, 2\}, \delta > 0, \label{eqn:minimax_zero_second_region_1} \\
        \text{If } x = x_V, & \text{ then } \
        \veta(x) = \begin{bmatrix}
            \frac{1}{2} + \frac{1}{2(\kappa + \delta)} \\
            \frac{1}{2} - \frac{1}{2(\kappa + \delta)} \\
            0 \\
            \vdots \\
            0
        \end{bmatrix}, \ 
        \widetilde{\veta}(x) = \ve_{1}, \label{eqn:minimax_zero_second_region_2}
    \end{align}
    where $\ve_i$ denotes the one-hot vector whose $i$-th element is one. 
    
    The triple  $(P_X, \veta, \widetilde{\veta})$ is thus parameterized by $j, \vb := [b_1 \ b_2 \ \cdots \ b_{V-1}]^\top$, and $\delta$.
    
    This construction ensures $(P_X, \veta, \widetilde{\veta}) \in \Pi(\epsilon, \kappa, V, 0)$. In particular,
    \begin{align*}
        & \mc{A}_\kappa \supseteq \{x_1, x_2, \dots, x_V \}, & P_X(\mc{A}_\kappa) \geq 1 - \epsilon, \\
        & \mc{X} \setminus \mc{A}_\kappa \subseteq \{x_0 \}, & P_X(\mc{X} \setminus \mc{A}_\kappa) \leq \epsilon,
    \end{align*}
    and $\widetilde{R}^* = 0$ because $\widetilde{\veta}(x)$ is one-hot for all $x$.
    

    For any classifier  $f$, 
    by definition, its risk equals
    \begin{align*}
        R \left( f \right) &= \mathbb{E}_{X,Y} \left[ \indicator{f(X)\ne Y} \right]\\
        &= \mathbb{E}_X \mathbb{E}_{Y|X}[\indicator{f(X)\ne Y}] \\
        &= \mathbb{E}_X \mathbb{E}_{Y|X}[1-\indicator{f(X)= Y}] \\
        &= \mathbb{E}_X \left[ 1 - [\veta(X)]_{f(X)} \right] \\
        & = \int_{\mc{X}} \left(1 - [\veta(x)]_{f(x)} \right) dP_X(x). 
    \end{align*}
    Therefore, the Bayes risk and excess risk equal
    \begin{align*}
        R(f^*) 
        & = \inf_f \ \mathbb{E}_{X,Y} \left[ \indicator{f(X)\ne Y} \right]\\
        & = \int_{\mc{X}} \left(1 - \max{} \veta(x) \right) dP_X(x), \\
        R(f) - R(f^*) 
        & = \int_{\mc{X}} \Big( \max \veta(x) - [\veta(x)]_{f(x)} \Big) \ dP_X(x).
    \end{align*}
    
    Under our construction of $P_{X}$, $R(f)$ can be decomposed into two parts
    \begin{align*}
        R \left( f \right) 
        & = \underbrace{ \int_{\{x_0 \}} \left(1 - [\veta(x)]_{f(x)} \right) dP_X(x)}_{:= R_0 (f)} + \underbrace{ \int_{\{x_1, \dots, x_V \}} \left(1 - [\veta(x)]_{f(x)} \right) dP_X(x) }_{:= R_V (f)},
    \end{align*}
    
    
    and so can the excess risk
    \begin{align*}
        R(f) - R(f^*) 
        & = \Big( R_{0} \left( f \right)  - R_{0} (f^*) \Big) + \Big( R_{V} \left( f \right)  - R_{V} (f^*) \Big) \\
        & = \int_{\{x_0\}} \Big( \max \veta(x) - [\veta(x)]_{f(x)} \Big) \ dP_X(x) \\
        & \qquad + \int_{\{x_1, \dots, x_V\}} \Big( \max \veta(x) - [\veta(x)]_{f(x)} \Big) \ dP_X(x).
    \end{align*}


    Recall that in our construction, $(P_X, \veta, \widetilde{\veta})$ is parameterized by $j, \vb$, and $\delta$.    
    Therefore
    \begin{align*}
        \sup_{ (P_X, \veta, \widetilde{\veta}) \in \Pi(\epsilon, \kappa, V, 0)} \expect{ Z^n }{ R \left( \hat{f} \right)  - R(f^*) } 
        & \geq \sup_{ j, \vb, \delta} \expect{ Z^n }{ R \left( \hat{f} \right)  - R(f^*) } \\
        & = \sup_{j, \vb, \delta} \Big\{ \expect{Z^n}{ R_{0} \left( \hat{f} \right)  - R_{0} (f^*) }  \\ 
        & \qquad +  \expect{Z^n}{ R_{V} \left( \hat{f} \right)  - R_{V} (f^*) } \Big\} \\
        & = \sup_{j} \ \expect{Z^n}{ R_{0} \left( \hat{f} \right)  - R_{0} (f^*) } \\ 
        & \qquad + \sup_{\vb, \delta} \ \expect{Z^n}{ R_{V} \left( \hat{f} \right)  - R_{V} (f^*) }
    \end{align*}
    where the last equality holds because region $\{x_0\}$ only depends on $j$, while region $\{x_1, \dots, x_V \}$ only depends on $\vb, \delta$.

    In the remaining part of the proof, we will examine 
    \begin{align}
        \sup_{j} \ \expect{Z^n}{ R_{0} \left( \hat{f} \right)  - R_{0} (f^*) } \label{eqn:minimax_zero_proof_first}
    \end{align}
    and 
    \begin{align}
        \sup_{\vb, \delta} \ \expect{Z^n}{ R_{V} \left( \hat{f} \right)  - R_{V} (f^*) } \label{eqn:minimax_zero_proof_second}
    \end{align}
    separately.

    Let's start with the first term \eqref{eqn:minimax_zero_proof_first}, which reflects the excess risk over the ``low signal strength'' region $\{ x_0 \}$. Since $\veta$ is one-hot on $\{ x_0 \}$, its Bayes risk over that is zero 
    \begin{align*}
        \sup_{j} \ \expect{Z^n}{ R_{0} \left( \hat{f} \right)  - R_{0} (f^*) } 
        & = \sup_{j} \ \expect{Z^n}{ R_{0} \left( \hat{f} \right) } \\
        & = \sup_{ j } \ \expect{ Z^n }{ \int_{\{x_0\}} \indicator{\hat{f}(x) \neq j} d P_X(x)}.
    \end{align*}
    To deal with $\sup_{j}$, we use a technique called ``the probabilistic method'' \citep{devroye1996probabilistic}: replace $j$ with a random variable $ J \sim \text{Uniform}\{1, 2, \dots, K\}$:
    \begin{align*}
        \sup_{ j } \ \expect{ Z^n }{ \int_{\{x_0\}} \indicator{\hat{f}(x) \neq j} d P_X(x)}  
        & \geq \expect{J}{ \expect{Z^n|J }{ \int_{\{x_0\}} \indicator{\hat{f}(x) \neq J} d P_X(x) }  } \\
        & = \expect{J,\ Z^n }{ \int_{\{x_0\}} \indicator{\hat{f}(x) \neq J} d P_X(x)} \\
        & = \expect{Z^n }{ \expect{J|Z^n }{ \int_{\{x_0\}} \indicator{\hat{f}(x) \neq J} d P_X(x) }  }.
    \end{align*}
    Again, notice that $J$ is an independent draw. Even if the point $x_0$ is observed in $Z^n$, the associated noisy label $\widetilde{Y} = 1$ does not give any information about the clean label $Y = J$. Thus
    \begin{align*}
        \expect{Z^n }{ \expect{J|Z^n }{ \int_{\{x_0\}} \indicator{\hat{f}(x) \neq J} d P_X(x) }  } 
        & = \expect{Z^n }{ \expect{J }{ \int_{\{x_0\}} \indicator{\hat{f}(x) \neq J} d P_X(x) }  } \\
        & = \expect{Z^n }{ \int_{\{x_0\}} \expect{J }{  \indicator{\hat{f}(x) \neq J}  } d P_X(x) } \\
        & = \expect{Z^n }{ \int_{\{x_0\}} \left(1 - \frac{1}{K} \right) d P_X(x) } \\
        & = \left(1 - \frac{1}{K} \right) \epsilon.
    \end{align*}
    
    Now we have the minimax lower bound for the first part \eqref{eqn:minimax_zero_proof_first}:
    \begin{align*}
        \sup_{j} \ \expect{Z^n}{ R_{\{x_0\}} \left( \hat{f} \right)  - R_{\{x_0\}} (f^*) } \geq \left(1 - \frac{1}{K} \right) \epsilon.
    \end{align*}

    For the second part \eqref{eqn:minimax_zero_proof_second}, 
    which is over $\{x_1, \dots, x_V\}$, due to the relative signal strength condition, and from our explicit construction in Eqn. \eqref{eqn:minimax_zero_second_region_1} and \eqref{eqn:minimax_zero_second_region_2}, the excess risks w.r.t. true and noisy distribution are related by
    \begin{align*}
        R_{V}(f) - R_{V}(f^*) 
        &= \int_{\{x_1, \dots, x_V\}} \Big( \max \veta(x) - [\veta(x)]_{f(x)} \Big) \ dP_X(x) \\
        &= \int_{\{x_1, \dots, x_V\}} \frac{1}{\kappa + \delta} \Big( \max \widetilde{\veta}(x) - [\widetilde{\veta}(x)]_{f(x)} \Big) \ dP_X(x) \quad \text{by construction of } \veta, \widetilde{\veta} \\
        & = \frac{1}{\kappa + \delta} \left( \widetilde{R}_{V}(f) - \widetilde{R}_{V}(\tilde{f}^*) \right),
    \end{align*}  
    where $\widetilde{R}_{V}(f) := \int_{\{x_1, \dots, x_V\}}  \Big( 1 - [\widetilde{\veta}(x)]_{f(x)} \Big) \ dP_X(x) $. 
    Also
    note that $f^*(x) = \tilde{f}^*(x) $ for $ x \in \{x_1, \dots, x_V\}$, which is a result of our construction of $\veta, \widetilde{\veta}$.
    
    Then 
    \begin{align*}
        \sup_{\vb, \delta} \ \expect{Z^n}{ R_{V} \left( \hat{f} \right)  - R_{V} (f^*) } = \sup_{\vb, \delta} \ \expect{Z^n}{ \frac{1}{\kappa + \delta} \left( \widetilde{R}_{V}(f) - \widetilde{R}_{V}(\tilde{f}^*) \right) }.
    \end{align*}
    This allows us to reduce the label noise problem to a standard learning problem: we have an iid sample $Z^n$ from $P_{X \widetilde{Y}}$ and consider the risk evaluated on the same distribution $P_{X \widetilde{Y}}$.
    The remainder of the proof is similar to the proof of 
    \citet[Theorem 14.1]{devroye1996probabilistic}.

    Notice that by our construction, $\widetilde{Y}$ is a deterministic function of $X$. To be specific, $\widetilde{Y} = \tilde{f}^*(X)$, where 
    \begin{align*}
        \tilde{f}^*(x) = \begin{cases}
            1 & x = x_0, \\
            b_t & x = x_t, \ 1 \leq t \leq V-1 \\
            1 & x = x_V
        \end{cases}
    \end{align*}
    is the noisy Bayes classifier.
    
    We use the shorthand $f_{\vb} := \tilde{f}^*$ to denote that the noisy Bayes classifier depends on $\vb$. 
    
    Since the noisy Bayes risk is zero, 
    \begin{align*}
        \sup_{\vb, \delta} \ \expect{Z^n}{ \frac{1}{\kappa + \delta} \left( \widetilde{R}_{V}(\hat{f}) - \widetilde{R}_{V}(\tilde{f}^*) \right) } 
        & = \sup_{\vb, \delta} \ \frac{1}{\kappa + \delta} \ \expect{Z^n}{  \widetilde{R}_{V}(\hat{f})  }.
    \end{align*}
    Again, use the probabilistic method, replace $\vb$ with $\B \sim \text{Uniform}\{1, 2\}^{V-1}$,
    \begin{align*}
        \sup_{\vb, \delta} \ \frac{1}{\kappa + \delta} \ \expect{Z^n}{  \widetilde{R}_{V}(\hat{f})  } 
        & \geq \sup_{\delta} \ \frac{1}{\kappa + \delta} \ \expect{\B, Z^n}{  \widetilde{R}_{V}(\hat{f})  } \\
        & = \sup_{\delta} \ \frac{1}{\kappa + \delta} \ \expect{Z^n }{ \expect{\B|Z^n }{ \int_{\{x_1, \dots, x_V \}} \indicator{\hat{f}(x) \neq f_{\B}(x)} d P_X(x) }  } \\
    \end{align*}

    

    Since we have $\B \sim \text{Uniform}\{1, 2\}^{V-1}$ and also $Z^n|\B \sim P_{X\widetilde{Y}}^{\otimes n} $, then by Bayes rule (or eye-balling, since $\widetilde{\veta}$ is one-hot), we get the posterior distribution of $\B | Z^n$, to be specific: $\forall x \in \{x_1, \cdots, x_V\},$
    \begin{align*}
        \text{If } x = X_i, i \in \left\{ 1, 2, \dots, n \right\}, & \quad \text{then } \ \prob{f_{\B}(x) = \widetilde{Y}_i | Z^n} = 1, \ \prob{f_{\B}(x) \neq \widetilde{Y}_i | Z^n} = 0 \\
        \text{If } x = x_V, & \quad \text{then } \ \prob{f_{\B}(x) = 1 | Z^n} = 1, \ \prob{f_{\B}(x) = 2 | Z^n} = 0 \\
        \text{If } x \notin \{X_1, \dots, X_n, x_V \}, & \quad \text{then } \ \prob{f_{\B}(x) = 1 | Z^n} = \half, \ \prob{f_{\B}(x) = 2 | Z^n} = \half ,
    \end{align*}
    where we overload the notation $\mathbb{P}$ to denote conditional probability of $\B | Z^n$.
    
    Then the optimal decision rule for predicting $\B$ based on sample $Z^n$ is: 
    \begin{align*}
        g(x; Z^n) 
        & = \begin{cases}
            \widetilde{Y}_i & x = X_i, i \in \left\{ 1, 2, \dots, n \right\} \\ 
            1 & x = x_V \\
            \text{random guess from } \{1, 2\} & x \neq X_1, \dots, x \neq X_n, x \neq x_V.
        \end{cases}
    \end{align*}
    Therefore, the error comes from the probability of $X \in \{x_1, \dots, x_V\}$ not being one of the observed $X_i$: for any $\hat{f}$,
    \begin{align*}
        \expect{\B, Z^n}{  \widetilde{R}_{V}(\hat{f})  }
        & = \expect{Z^n }{ \expect{\B|Z^n }{ \int_{\{x_1, \dots, x_V \}} \indicator{\hat{f}(x) \neq f_{\B}(x)} d P_X(x) }  } \\
        & \geq \prob{  X \in \{x_1, \dots, x_V \}, \  g(X; Z^n) \neq f_{\B}(X) } \qquad \because \text{error of $\hat{f} \ge $  error of $g$} \\
        & = \left(1 - \frac{1}{2} \right) \prob{X \neq X_1, \dots, X \neq X_n, X \neq x_V, X \in \{x_1, \dots, x_V\}} \\
        & =  \frac{1}{2}  \sum_{t=1}^{V} \prob{X \neq X_1, \dots, X \neq X_n, X \neq x_V, X = x_t} \\
        & =  \frac{1}{2}  \sum_{t=1}^{V} \prob{X_1 \neq x_t, \dots, X_n \neq x_t, x_V \neq x_t, X = x_t} \quad \because \text{replace all $X$ with $x_t$} \\
        & =  \frac{1}{2}  \sum_{t=1}^{V-1} \prob{X_1 \neq x_t, \dots, X_n \neq x_t, X = x_t} \\
        & = \frac{1}{2} \sum_{t=1}^{V-1} \prob{X_1 \neq x_t, \dots, X_n \neq x_t | X = x_t} \prob{X = x_t} \\
        & = \frac{1}{2} \sum_{t=1}^{V-1} \left(1 - \prob{X = x_t} \right)^n \prob{X = x_t} \\
        & = \frac{1}{2}  (V-1) \left(1 - \frac{1-\epsilon}{n} \right)^n  \left( \frac{1-\epsilon}{n} \right)\\
        & = \frac{(V-1)(1-\epsilon)}{2n} \left(1 - \frac{1-\epsilon}{n} \right)^n \\
        & = \frac{(V-1)(1-\epsilon)}{2n} \left(1 - \frac{1-\epsilon}{n} \right)^{1 + \epsilon} \left(1 - \frac{1-\epsilon}{n} \right)^{n - 1 - \epsilon} \\
        & \geq \frac{(V-1)(1-\epsilon)}{2n} \left(1 - \frac{1-\epsilon}{n} \right)^{1 + \epsilon} e^{-1+\epsilon} \qquad \because \left(1 - \frac{1-\epsilon}{n} \right)^{n-1-\epsilon} \downarrow e^{-1+\epsilon} \\
        & \geq \frac{(V-1)(1-\epsilon)}{2n} \left(1 - \frac{1}{n} \right)^{2} e^{-1} \qquad \qquad  \because \epsilon \in [0, 1] \\
        & \geq \frac{(V-1)(1-\epsilon)}{2n}  \frac{e^{-1}}{4} = \frac{(V-1)(1-\epsilon)}{8en} \qquad \text{take } n > 2.
    \end{align*}
    Now we get the minimax risk for the second part \eqref{eqn:minimax_zero_proof_second}
    \begin{align*}
        \sup_{\vb, \delta} \ \expect{Z^n}{ R_{\mc{A}_{\kappa}} \left( \hat{f} \right)  - R_{\mc{A}_{\kappa}} (f^*) } 
        & \geq \sup_{\delta} \ \frac{1}{\kappa + \delta} \frac{(V-1)(1-\epsilon)}{8en} \\
        & \geq \frac{1}{\kappa} \frac{(V-1)(1-\epsilon)}{8en} \qquad \text{let } \delta \downarrow 0
    \end{align*}

    Combine the two parts together, we get the final result, for $n > \max (V - 1, 2)$
    \begin{align*}
        \sup_{ (P_X, \veta, \widetilde{\veta}) \in \Pi(\epsilon, \kappa, V, 0)} \expect{ Z^n }{ R \left( \hat{f} \right)  - R(f^*) }
        & \geq
        \frac{K-1}{K}  \epsilon + \frac{1}{\kappa}  \frac{(V-1)(1-\epsilon)}{8en} .
    \end{align*}
    

\end{proof}

As for the general version of the lower bound,
now consider the set of triples:
\begin{align*}
    \Pi(\epsilon, \kappa, V, L) 
    := \Big\{ 
    \left( P_X, \veta, \widetilde{\veta} \right): &
    P_X \Big( \mc{A}_{\kappa} \left(\veta, \widetilde{\veta} \right) \Big) \geq 1 - \epsilon, \\
        & P_X \text{ supported on $V+1$ points},  \frac{ \widetilde{R}_{\mc{A}_{\kappa}} \left(\tilde{f}^* \right)}{P_X\Big( \mc{A}_{\kappa} \left(\veta, \widetilde{\veta} \right) \Big)} \leq L
    \Big\},
\end{align*}
where $\widetilde{R}_{C}(f) = \int_C \left( 1 - [\widetilde{\veta}(x)]_{f(x)} \right) dP_X(x)$.

\begin{theorem*}[Minimax Lower Bound (General Version)]
    Let $\epsilon \in [0, 1], \kappa > 0, V > 1, L \in (0, 1/2)$.
    For any learning rule $\hat{f}$ based upon $ Z^n = \big\{ ( X_i, \widetilde{Y}_i ) \big\}_{i=1}^n $, for $n \geq \frac{V-1}{2L} \max \left\{16, \frac{1}{(1-2L)^2} \right\}$
    \begin{align*}
        \sup_{ (P_X, \veta, \widetilde{\veta}) \in \Pi(\epsilon, \kappa) } \expect{ Z^n }{ R \left( \hat{f} \right)  - R(f^*) } 
        & \geq \sup_{ (P_X, \veta, \widetilde{\veta}) \in \Pi(\epsilon, \kappa, V, L) } \expect{ Z^n }{ R \left( \hat{f} \right)  - R(f^*) } \\
        & \geq \frac{K-1}{K}  \epsilon 
        + \frac{1-\epsilon}{\kappa} \sqrt{\frac{(V-1)L}{2n}} e^{-7}
        \\
        & = \frac{K-1}{K}  \epsilon + \Omega \left( \frac{1}{\kappa} \ \sqrt{\frac{1}{n}} \right).
    \end{align*}
\end{theorem*}

\begin{proof}

Now we construct a triple  $(P_X, \veta, \widetilde{\veta})$ that is parameterized by $j, \vb := [b_1 \ b_2 \ \cdots \ b_{V-1}]^\top$, $\delta$, $c$ and $p$. 

First, we define $P_X$. Pick any $V+1$ distinct points $x_0, x_1, \ldots, x_V$,
\begin{align*}
    P_X(x) = \begin{cases}
        \epsilon & x = x_0 \\
        (1 - \epsilon) \cdot p & x = x_1, \dots, x_{V-1} \\
        (1 - \epsilon) \cdot \left(1 - (V-1)p \right) & x = x_V.
    \end{cases}
\end{align*}
This imposes the constraint $(V-1)p \leq 1$, which will be satisfied in the end. Notice the difference compared to the previous zero-error proof: we place probability mass $p$, rather than $1/n$, on $x_1, \ldots, x_{V-1}$.

As for the clean and noisy class probabilities, choose
\begin{align}
    \text{If } x = x_0, & \text{ then } \ \veta(x) = \ve_j, \ \widetilde{\veta}(x) = \ve_1, \quad j \in \{1, 2, \dots k \} \label{eqn:minimax_first_region} \\
    \text{If } x = x_t, & \ 1 \leq t \leq V-1,  \text{ then } \ 
    \veta(x) = \begin{bmatrix}
        \frac{1}{2} + \frac{c}{\kappa + \delta} \cdot (-1)^{b_t + 1} \\
        \frac{1}{2} - \frac{c}{\kappa + \delta} \cdot (-1)^{b_t + 1} \\
        0 \\
        \vdots \\
        0
    \end{bmatrix}, \ 
    \widetilde{\veta}(x) = \begin{bmatrix}
            \frac{1}{2} + c \cdot (-1)^{b_t + 1} \\ 
            \frac{1}{2} - c \cdot (-1)^{b_t + 1} \\ 
            0 \\ 
            \vdots \\
            0
        \end{bmatrix}, \nonumber  \\
    & \qquad \qquad \qquad b_t \in \{1, 2\}, \ \delta > 0, \ c \in \left(0, \frac{1}{2} \right) \label{eqn:minimax_second_region_1}  \\
    \text{If } x = x_V, & \text{ then } \ 
    \veta(x) = \begin{bmatrix}
        \frac{1}{2} + \frac{1}{2(\kappa + \delta)} \\
        \frac{1}{2} - \frac{1}{2(\kappa + \delta)} \\
        0 \\
        \vdots \\
        0
    \end{bmatrix}, \ 
    \widetilde{\veta}(x) = \ve_{1}, \label{eqn:minimax_second_region_2}
\end{align}
where $\ve_i$ denotes the one-hot vector whose $i$-th element is one.

The construction for class posterior is also similar to the previous proof, except that for $x = x_t, t \in \{1, \ldots, V-1\}$, $\widetilde{\veta}$ is no longer a one-hot vector, rather has class probability separated by $2c$: $ \Big| \left[\widetilde{\veta}(x)\right]_1 - \left[\widetilde{\veta}(x)\right]_2 \Big| = 2c$.

Therefore, the triple  $(P_X, \veta, \widetilde{\veta})$ can be parameterized by $j, \vb := [b_1 \ b_2 \ \cdots \ b_{V-1}]^\top$, $\delta$, $c$ and $p$.

Again, this construction ensures $(P_X, \veta, \widetilde{\veta}) \in \Pi(\epsilon, \kappa)$, to be specific:
    \begin{align*}
        & \mc{A}_\kappa \supseteq \{x_1, x_2, \dots, x_V \}, & P_X(\mc{A}_\kappa) \geq 1 - \epsilon, \\
        & \mc{X} \setminus \mc{A}_\kappa \subseteq \{x_0 \}, & P_X(\mc{X} \setminus \mc{A}_\kappa) \leq \epsilon.
    \end{align*}

For any classifier $f$, its risk can be decomposed into two parts
    \begin{align*}
        R \left( f \right) 
        & = \underbrace{ \int_{\{x_0 \}} \left(1 - [\veta(x)]_{f(x)} \right) dP_X(x)}_{:= R_0 (f)} + \underbrace{ \int_{\{x_1, \dots, x_V \}} \left(1 - [\veta(x)]_{f(x)} \right) dP_X(x) }_{:= R_V (f)},
    \end{align*}

as can its excess risk
\begin{align*}
    R \left( f \right)  - R(f^*) = \Big( R_{0} \left( f \right)  - R_{0} (f^*) \Big) + \left( R_{V} \Big( f \right)  - R_{V} (f^*) \Big).
\end{align*}

In our construction, $(P_X, \veta, \widetilde{\veta})$ is parameterized by $j, \vb := [b_1 \ b_2 \ \cdots \ b_{V-1}]^\top$, $\delta$, $c$ and $p$, therefore 
\begin{align}
    \sup_{ (P_X, \veta, \widetilde{\veta}) \in \Pi(\epsilon, \kappa, V, L)} \expect{ Z^n }{ R \left( \hat{f} \right)  - R(f^*) } &
    \geq \sup_{j} \ \expect{Z^n}{ R_{0} \left( \hat{f} \right)  - R_{0} (f^*) } \label{eqn:minimax_proof_first} \\ 
    &  \quad + \sup_{\vb, \delta, c, p} \ \expect{Z^n}{ \frac{1}{\kappa + \delta} \left( \widetilde{R}_{V}(f) - \widetilde{R}_{V}(\tilde{f}^*) \right) } \label{eqn:minimax_proof_second}.
\end{align}

Note that we have used the fact that 
\begin{align*}
    R_{V}(f) - R_{V}(f^*) 
    & = \frac{1}{\kappa + \delta} \left( \widetilde{R}_{V}(f) - \widetilde{R}_{V}(\tilde{f}^*) \right),
\end{align*}  
where $\widetilde{R}_{V}(f) := \int_{\{x_1, \dots, x_V\}}  \Big( 1 - [\widetilde{\veta}(x)]_{f(x)} \Big) \ dP_X(x) $. 

The first part \eqref{eqn:minimax_proof_first} is exactly the same as in the zero-error proof, and we have 
\begin{align*}
    \sup_{j} \ \expect{Z^n}{ R_{0} \left( \hat{f} \right)  - R_{0} (f^*) } \geq \left(1 - \frac{1}{K} \right) \epsilon.
\end{align*}

From this point forward, the procedure is similar to the proof of  \citet[][Theorem 14.5]{devroye1996probabilistic}.
For the second part \eqref{eqn:minimax_proof_second}, the noisy Bayes classifier is still
\begin{align*}
        \tilde{f}^*(x) = \begin{cases}
            j & x = x_0, \\
            b_t & x = x_t, \ 1 \leq t \leq V \\
            1 & x = x_V.
        \end{cases}
\end{align*}
We also use the shorthand $f_{\vb} := \tilde{f}^*$ to denote that the noisy Bayes classifier depends on $\vb$. 

Now the noisy Bayes risk is no longer zero. 
In fact
\begin{align*}
    \widetilde{R}_{V}(\tilde{f}^*) = \int_{\{x_1, \dots, x_V \}} \left(1 - [\widetilde{\veta}(x)]_{f(x)} \right) dP_X(x) = (V-1) (1 - \epsilon) p \left(\frac{1}{2} - c \right) 
\end{align*}

What's more, 
\begin{equation}
     \frac{ \widetilde{R}_{\mc{A}_{\kappa}} \left(\tilde{f}^* \right)}{P_X\Big( \mc{A}_{\kappa} \left(\veta, \widetilde{\veta} \right) \Big)} \leq \frac{ \widetilde{R}_{V}(\tilde{f}^*) }{P_X\Big( \left\{x_1, \dots, x_V \right\} \Big)} = (V-1)p \left(\frac{1}{2} - c \right),
        \label{eqn:lpc_relation}
\end{equation}
where the inequality holds from $ \widetilde{R}_{\mc{A}_{\kappa}}(\tilde{f}^*) = \widetilde{R}_{V}(\tilde{f}^*) $ (because $\widetilde{\veta}$ is one-hot at point $x_0$) and $P_X\Big( \mc{A}_{\kappa} \left(\veta, \widetilde{\veta} \right) \Big) \geq P_X\Big( \left\{x_1, \dots, x_V \right\} \Big)$.


Notice that in order to ensure that our construction $(P_X,\veta, \tveta) \in \Pi(\epsilon, \kappa, V,L)$, by definition 
\begin{equation*}
        \frac{ \widetilde{R}_{\mc{A}_{\kappa}} \left(\tilde{f}^* \right)}{P_X\Big( \mc{A}_{\kappa} \left(\veta, \widetilde{\veta} \right) \Big)} 
        \leq L,
\end{equation*}

Due to the upper bound of \eqref{eqn:lpc_relation}, it suffices to require that
\begin{equation}
    (V-1)p \left(\frac{1}{2} - c \right) = L,
    \label{eqn:requirement_on_L}
\end{equation}

and this ensures that $(P_X , \veta, \tveta) \in \Pi(\epsilon, \kappa, V,L)$ upon recalling that $(P_X, \veta, \tveta) \in \Pi(\epsilon, \kappa),$ and that $P_X$ is supported on $V+1$ points.



It should be noted that since $(V-1)p \leq 1$ is required, and since $c > 0$, we have $L < 1 \cdot 1/2$.
This is the origin of our condition $L < 1/2$ in the statement of the theorem. Naturally, the statement can be adjusted to $\min(L,1/2)$ instead. In any case, we are left with two nontrivial constraint on our parameters $(p,c)$: (\ref{eqn:requirement_on_L}) and $(V-1) p \le 1$, along with the boundary consraints $p \in [0,1]$ and $c \in [0,1/2]$.

For fixed $\vb$, plugging in the definition of $\widetilde{\veta}$, the excess risk over region $\left\{x_1, \dots, x_V \right\}$ becomes
\begin{align*}
    \widetilde{R}_{V}(\hat{f}) - \widetilde{R}_{V}(\tilde{f}^*)
    & = \int_{\{x_1, \dots, x_V \}} 2c \indicator{\hat{f}(x) \neq f_{\vb}(x)} d P_X(x) \\
    & \geq 2 c \sum_{t=1}^{V-1}  (1 - \epsilon) p \indicator{ \hat{f}(x_t) \neq f_{\vb}(x_t)},
\end{align*}
where the inequality follows from the fact that we ignore the risk on point $x_V$.

Using the probabilistic method, replace $\vb$ with $\B \sim \text{Uniform}\{1, 2\}^{V-1}$,
\begin{align*}
    \sup_{\vb, \delta, c, p} \ \expect{Z^n}{ \frac{1}{\kappa + \delta} \left( \widetilde{R}_{V}(\hat{f}) - \widetilde{R}_{V}(\tilde{f}^*) \right) } 
    & \geq \sup_{\delta, c, p} \ \expect{\B, Z^n}{ \frac{1}{\kappa + \delta} \left( \widetilde{R}_{V}(\hat{f}) - \widetilde{R}_{V}(\tilde{f}^*) \right) } \\
    & = \sup_{\delta, c, p} \frac{1}{\kappa + \delta} \expect{Z^n}{ \expect{\B | Z^n}{ \left( \widetilde{R}_{V}(\hat{f}) - \widetilde{R}_{V}(\tilde{f}^*) \right) } } 
\end{align*}


Now, we need to calculate $\B |Z^n$, which can be calculated using Bayes rule because we have $\B \sim \text{Uniform}\{1, 2\}^{V-1}$ and also $Z^n|\B \sim P_{X\widetilde{Y}}^{\otimes n}$. 

To be specific, for any $x \in \{x_0, x_1, \dots, x_{V-1} \}$, assume point $x_t$ is observed $k$ times in training sample $Z^n$,
\begin{align*}
    \prob{f_{\B}(x) = 1 | Z^n} = \begin{cases}
        \frac{1}{2} & x \neq X_1, \dots, x \neq X_n, x \neq x_V \\
        \prob{B_t = 1 | Y_{t_1}, \dots, Y_{t_k}} & x = x_t = X_{t_1} = \cdots = X_{t_k}, \ 1 \leq t \leq V-1,
    \end{cases}
\end{align*}
where $B_t$ denotes the $t$-th element of vector $\B$ (that associates with $x_t$).

Next we compute $\prob{B_t = 1 | Y_{t_1} = y_1, \dots, Y_{t_k} = y_k}$ for $y_1, \dots, y_k \in \{1, 2\}$. Denote the numbers of ones and twos by $k_1 = \abs{\{j \leq k: y_j = 1 \}}$ and $k_2 = \abs{\{j \leq k: y_j = 2 \}}$. Using Bayes rule, we get 
\begin{align*}
    \prob{B_t = 1 | Y_{t_1}, \dots, Y_{t_k}} 
    & = \frac{ \prob{B_t = 1 \cap Y_{t_1}, \dots, Y_{t_k}} }{ \prob{ Y_{t_1}, \dots, Y_{t_k}} } \\
    & = \frac{ \prob{Y_{t_1}, \dots, Y_{t_k} | B_t = 1} \prob{B_t =1} }{ \sum_{i = 1}^2 \prob{Y_{t_1}, \dots, Y_{t_k} | B_t = i} \prob{B_t = i}  } \\
    & = \frac{ (1/2+c)^{k_1} (1/2 - c)^{k_2} (1/2) }{ 
    (1/2+c)^{k_1} (1/2 - c)^{k_2} (1/2) + (1/2+c)^{k_2} (1/2 - c)^{k_1} (1/2)}.
\end{align*}

After some calculation, following the proof of  \citet[][Theorem 14.5]{devroye1996probabilistic}, we get 
\begin{align}
    \sup_{\vb, \delta, c, p} \ & \expect{Z^n}{ \frac{1}{\kappa + \delta}  \left( \widetilde{R}_{\mc{A}_{\kappa}}(f) - \widetilde{R}_{\mc{A}_{\kappa}}(\tilde{f}^*) \right) } \notag \\
    & \qquad \geq \sup_{\delta, c, p} \ \frac{1}{\kappa + \delta} c(V-1)(1-\epsilon)p e^{- \frac{8n(1-\epsilon)pc^2}{1-2c} - \frac{4 c \sqrt{n(1-\epsilon)p}}{1-2c}} \notag\\
    & \qquad \geq \frac{1-\epsilon}{\kappa} \sup_{c, p} c (V-1) p e^{- \frac{8npc^2}{1-2c} - \frac{4 c \sqrt{np}}{1-2c}} \quad \because \epsilon \geq 0, \text{ take } \delta \downarrow 0 \notag\\
    & \qquad = \frac{1-\epsilon}{\kappa} \sup_{c, p} c \ \frac{L}{1/2 - c} e^{- \frac{8npc^2}{1-2c} - \frac{4 c \sqrt{np}}{1-2c}}, \qquad \because  \eqref{eqn:requirement_on_L}  \label{eqn:lower_bound_pre_optimisation}
\end{align}

where the supremum is over $(p, c) \in [0,1] \times [0,1/2]$ such that 
\[ (V-1) p \le 1, \text{ and } (V-1)p (1/2 - c) = L.\] Now, suppose $n$ is so large that 
\[ n \ge \frac{(V-1)}{8L(1/2 - L)^2} \iff L  \le \frac12 -  \sqrt{\frac{(V-1)}{8n L}}, \] 
and further that 
\[ \sqrt{\frac{(V-1)}{8nL}} \le \frac18 \iff n \ge \frac{8(V-1)}{L}. \] 
We choose 
\[ c= \sqrt{\frac{(V-1)}{8nL}}, \textrm{ and } p = \frac{L}{(V-1)(1/2 - c)}.\] By our choice of $c$ and the first condition on $n$ above, we can conclude that $L \le (1/2 - c)$, and therefore, \[(V-1) p = \frac{L}{1/2 - c} \le 1, \] meaning that both the constraints required on $(p,c)$ are met by the above choice. 

As a consequence of this choice of $c,p$, we observe that \[ n pc^2 = \frac{n L}{(V-1)(1/2 - c)} \cdot c^2 = \frac{nL}{(V-1)(1/2-c) } \cdot \frac{(V-1)}{8nL} = \frac{1}{4-8c} \le \frac13.   \]

Since $c \le 1/8$ further implies that $\frac{1}{1-2c} \le \frac{4}{3},$ this implies that \[\frac{8npc^2}{1-2c} + \frac{4 \sqrt{npc^2}}{1-2c} \le \frac{8}{3} \cdot \frac{4}{3} + 4 \cdot \frac{4}{3} \cdot \sqrt{\frac{1}{3}} \le 7.\]

Thus, instantiating the bound (\ref{eqn:lower_bound_pre_optimisation}), we conclude that \begin{align*}
    \sup_{\vb, \delta, c, p} \ \expect{Z^n}{ \frac{1}{\kappa + \delta} \left( \widetilde{R}_{\mc{A}_{\kappa}}(f) - \widetilde{R}_{\mc{A}_{\kappa}}(\tilde{f}^*) \right) } 
    &\geq \frac{1-\epsilon}{\kappa} \cdot  \sqrt{\frac{V-1}{8nL}} \cdot  \frac{L}{1/2 - c} \cdot  e^{-7} \\ 
    &\ge \frac{1-\epsilon}{\kappa} \sqrt{\frac{(V-1)L}{8n}} e^{-7}\cdot 2 \\
    &=\frac{1-\epsilon}{\kappa} \sqrt{\frac{(V-1)L}{2n}} e^{-7} .
\end{align*}

Putting the two parts together
\begin{align*}
    \sup_{ (P_X, \veta, \widetilde{\veta}) \in \Pi(\epsilon, \kappa) } \expect{ Z^n }{ R \left( \hat{f} \right)  - R(f^*) } 
    & \geq \frac{K-1}{K}  \epsilon + \frac{1-\epsilon}{\kappa} \sqrt{\frac{(V-1)L}{2n}} e^{-7}, 
\end{align*}
for $n \geq \frac{V-1}{2L} \max \left\{16, \frac{1}{(1-2L)^2} \right\}$.

\end{proof}

\subsubsection{Proof of upper bound: Lemma \ref{lem:excess_risk}}
\label{subsec:oracle_proof}

\begin{lemma*}[Oracle Inequality under Feature-dependent Label Noise]
    For any $(P_X, \veta, \widetilde{\veta}) $ and any classifier $f$, we have
    \begin{align*}
        R(f) - R(f^*) \leq \inf_{\kappa>0} \left\{ P_X \Big( \mc{X} \setminus \mc{A}_{\kappa} \left(\veta, \widetilde{\veta} \right) \Big) + \frac{1}{\kappa} \left( \widetilde{R}(f) - \widetilde{R} \left( \tilde{f}^* \right) \right) \right\} .
    \end{align*}
\end{lemma*}

\begin{proof}
    For any $\kappa \geq 0$, the input space $\mc{X}$ can be divided into two regions: $\mc{X} \setminus \mc{A}_{\kappa}$ and $\mc{A}_{\kappa}$.

    
    For any $f$, its risk is
    \begin{align*}
        R \left( f \right) &= \mathbb{E}_{X,Y} \left[ \indicator{f(X)\ne Y} \right]\\
        &= \mathbb{E}_X \mathbb{E}_{Y|X}[\indicator{f(X)\ne Y}] \\
        &= \mathbb{E}_X \mathbb{E}_{Y|X}[1-\indicator{f(X)= Y}] \\
        &= \mathbb{E}_X \left[ 1 - [\veta(X)]_{f(X)} \right] \\
        & = \int_{\mc{X}} \left(1 - [\veta(x)]_{f(x)} \right) dP_X(x). 
    \end{align*}

    Therefore, its excess risk is
    \begin{align*}
        R(f) - R(f^*) 
        & = \int_{\mathcal{X}} \Big( \max \veta(x) - [\veta(x)]_{f(x)} \Big) \ dP_X(x) \\
        & = 
        \underbrace{\int_{\mc{X} \setminus \mc{A}_{\kappa}} \Big( \max \veta(x) - [\veta(x)]_{f(x)} \Big) \ dP_X(x)}_{\textcircled{a}} \\
        & \qquad + \underbrace{\int_{ \mc{A}_{\kappa}} \Big( \max \veta(x) - [\veta(x)]_{f(x)} \Big) \ dP_X(x)}_{\textcircled{b}} 
    \end{align*}

    Now examine the two terms separately,
    \begin{align*}
        \textcircled{a} & \leq \int_{\mc{X} \setminus \mc{A}_{\kappa}} 1 \ dP_X(x) = P_X \Big( \mc{X} \setminus \mc{A}_{\kappa} \left(\veta, \widetilde{\veta} \right) \Big),
    \end{align*}
    and
    \begin{align*}
        \textcircled{b} 
        & < \int_{\mc{A}_{\kappa}} \frac{1}{\kappa} \Big( \max \widetilde{\veta}(x) - [\widetilde{\veta}(x)]_{f(x)} \Big) \ dP_X(x) & \because \text{by definition of relative signal strength} \\ 
        & \leq \int_{\mc{X}} \frac{1}{\kappa} \Big( \max \widetilde{\veta}(x) - [\widetilde{\veta}(x)]_{f(x)} \Big) \ dP_X(x) \\
        & = \frac{1}{\kappa} \left( \widetilde{R}(f) - \widetilde{R}(\tilde{f}^*) \right) & \because \text{by definition of $\widetilde{R}$}.
    \end{align*}
    Since this works for any $\kappa > 0$, we then have 
    \begin{align*}
        R(f) - R(f^*) \leq \inf_{\kappa > 0} \left\{ P_X \Big( \mc{X} \setminus \mc{A}_{\kappa} \left(\veta, \widetilde{\veta} \right) \Big) + \frac{1}{\kappa} \left( \widetilde{R}(f) - \widetilde{R} \left( \tilde{f}^* \right) \right) \right\} .
    \end{align*}

\end{proof}

\subsubsection{Proof of upper bound: \Cref{thm:erm_generalization}}
\label{subsec:erm_generalization_proof}

To set the stage for the rate of convergence proof, we first introduce the concept of shattering in the multiclass setting and the Natarajan dimension \citep{natarajan1989learning}, which serves as a multiclass counterpart to the VC dimension \citep{vapnik1971}.




    
    

\begin{definition}[Multiclass Shattering] Let $\mc{H}$ be a class of functions from $\mc{X}$ to $\mc{Y} = \{1,2, \dots, K \}$. For any set containing $n$ distinct elements $C_n = \{x_1, \dots, x_n \} \subset \mc{X}$, 
denote 
\begin{align*}
    \mc{H}_{C_n} = \left\{ \left(h(x_1), \dots, h(x_n) \right): h \in \mc{H} \right\},
\end{align*}
and therefore $\abs{\mc{H}_{C_n} }$
is the number of distinct vectors of length $n$ that can be realized by functions in $\mc{H}$.

The \emph{$n^{th}$ shatter coefficient} is defined as 
\begin{align*}
    S(\mc{H}, n) := \max_{C_n} \abs{\mc{H}_{C_n} }.
\end{align*}

We say that a set $C_n$ is shattered by $\mc{H}$ if there exists $f, g: C_n \rightarrow \mc{Y}$ such that for every $ x \in C_n$, $f(x) \neq g(x)$, and 
\begin{align*}
    \mc{H}_C \supseteq \{f(x_1), g(x_1) \} \times \{f(x_2), g(x_2) \} \times \cdots \times \{f(x_n), g(x_n) \}
\end{align*}
    
\end{definition}

If $\mc{Y} = \{1, 2\}$, this definition reduces to the binary notion of shattering which says all labeling of points can be realized by some function in the hypothesis class $\mc{H}$, i.e., $\mc{H}_C = \{1, 2\}^{\abs{C}}$.
Note that multiclass shattering does not mean being able to realize all $K$ possible labels for each point $x \in C$. Instead, multiclass shattering is more like ``embed the binary cube into multiclass'', where
every $x \in C$ is allowed to pick from two of the $K$ labels.

\begin{definition}[Natarajan Dimension]
    The Natarajan dimension of $\mc{H}$, denoted Ndim($\mc{H}$), is the maximal size of a shattered set $C \in \mc{X}$.
\end{definition}

\begin{theorem*}[Excess Risk Upper Bound of NI-ERM] 
    Let $\epsilon \in [0, 1] , \kappa \in (0, + \infty)$.
    Consider any $(P_X, \veta, \widetilde{\veta}) \in \Pi(\epsilon, \kappa)$, assume function class $\mathcal{F}$ has Natarajan dimension $V$, and the noisy Bayes classifier $\tilde{f}^*$ belongs to $\mathcal{F}$.
    Let $\hat{f} \in \mathcal{F}$ be the ERM trained on $ Z^n = \big\{ ( X_i, \widetilde{Y}_i ) \big\}_{i=1}^n $, then
    \begin{align*}
        \expect{ Z^n }{ R \left( \hat{f} \right)  - R(f^*) } 
        & \leq \epsilon + \frac{1}{\kappa} \cdot 16 \sqrt{ \frac{V \log n + 2V \log k + 4}{2n} } \\
        & = \epsilon + \mc{O} \left( \frac{1}{\kappa} \sqrt{\frac{V}{n}} \right) \quad \text{up to log factor} .
    \end{align*}
\end{theorem*}

\begin{proof}
    Following directly from Lemma \ref{lem:excess_risk}, with $(P_X, \veta, \widetilde{\veta}) \in \Pi(\epsilon, \kappa)$, we already have
    \begin{align*}
        R(f) - R(f^*) & \leq  P_X \Big( \mc{X} \setminus \mc{A}_{\kappa} \left(\veta, \widetilde{\veta} \right) \Big) + \frac{1}{\kappa} \left( \widetilde{R}(f) - \widetilde{R} \left( \tilde{f}^* \right) \right) \\ 
        & \leq \epsilon + \frac{1}{\kappa} \left( \widetilde{R}(f) - \widetilde{R} \left( \tilde{f}^* \right) \right).
    \end{align*}
    Now replace $f$ with NI-ERM$\hat{f}$. To bound the expected excess risk we employ a multiclass VC-style inequality. 

    \begin{lemma}
    \label{lem:dgl:12.1}
    \begin{align*}
        \expect{ Z^n }{ \widetilde{R} \left( \hat{f} \right) - \widetilde{R} \left( \tilde{f}^* \right) } \leq 16 \sqrt{ \frac{ \log (8e S(\mc{H}, n)) }{2n} }
    \end{align*}
    \label{lemma:shattering}
    \end{lemma}

The binary version of this lemma is Corollary 12.1 in \citet{devroye1996probabilistic}. We prove the multiclass version below in Section \ref{subsec:mcshatter_proof}.

    
        
        


Next, we bound the multiclass shattering coefficient with Natarajan dimension, using the following  lemma, which can be viewed as a multiclass version of Sauer's lemma.
    
    \begin{lemma}[\citet{natarajan1989learning}]
        Let $C$ and $\mc{Y}$ be two finite sets and let $\mc{H}$ be a set of functions from $C$ to $\mc{Y}$. Then
        \begin{align*}
            \abs{\mc{H}} \leq \abs{C }^{\text{Ndim}(\mc{H})} \cdot \abs{\mc{Y}}^{2\text{Ndim}(\mc{H})}.
        \end{align*}
    \end{lemma}
    


Letting $V$ denote $\text{Ndim}(\mc{H})$, we have that $S(\mc{H}, n) \le n^V K^{2V}$, and therefore \Cref{lemma:shattering} can be upper bounded by
    \begin{align*}
        \expect{ Z^n }{ \widetilde{R} \left( \hat{f} \right) - \widetilde{R} \left( \tilde{f}^* \right) }  
        & \leq 16 \sqrt{ \frac{ \log \left( 8 e (n)^V K^{2V} \right) }{2n} } \\
        & = \ 16 \sqrt{ \frac{ \log 8e +  \log \left(n^V \right) + \log 
        \left( K^{2V} \right) }{2n} } \\
        & \leq 16 \sqrt{ \frac{V \log n + 2V \log K + 4}{2n} } \\
    \end{align*}
    Putting things together, 
    \begin{align*}
        \expect{ Z^n }{ R \left( \hat{f} \right)  - R(f^*) } 
        \leq \epsilon + 
        \frac{1}{\kappa} \cdot 16 \sqrt{ \frac{V \log n + 2V \log K + 4}{2n} }.
    \end{align*}  
\end{proof}

\subsubsection{Proof of Lemma \ref{lem:dgl:12.1}}
\label{subsec:mcshatter_proof}

\begin{theorem}
\label{thm:mcshatter}
Consider any set of multiclass classifiers $\mc{F}$. Let $(X_1,Y_1),\ldots,(X_n,Y_n)$ be iid draws from $P_{XY}$. For any $n$, and any $\epsilon > 0$,
$$
\Pr\left\{ \sup_{f \in \mc{F}} |R_n(f) - R(f)| > \epsilon \right\}
\le 8S(\mc{F},n)e^{-n \epsilon^2/32}
$$
where the probability is with respect to the draw of the data.
\end{theorem}

\begin{proof}
Apply Theorem 12.5 from \citet{devroye1996probabilistic}, with the following identifications. In what follows, the left-hand side of each equation is a notation from \citet{devroye1996probabilistic}, and the right-hand side is our notation.
\begin{align*}
\nu &= P_{XY} \\
Z &= (X,Y) \\
Z_i &= (X_i,Y_i) \\
\mc{A} &= \{A_f \mid f \in \mc{F}\}, \text{ where } A_f := \{(x,y) \mid f(x) = y\}
\end{align*}
With these identifications, we have
\begin{align*}
\nu(A_f) &= 1 - R(f) \\
\nu_n(A_f) &= \frac1{n} \sum_i \indicator{\{ Z_i \in A_f \}} = \frac1{n} \sum_i \indicator{\{ f(X_i) = Y_i\}} = 1 - R_n(f) 
\end{align*}
By Theorem 12.5 we conclude
$$
\Pr\left\{ \sup_{f \in \mc{F}} |R_n(f) - R(f)| > \epsilon \right\}
\le 8s(\mc{A},n)e^{-n \epsilon^2/32},
$$
where $s(\mc{A},n)$ (note the lowercase ``s'') is defined to be 
$$
\max_{z_1,\ldots,z_n} \mc{N}_{\mc{A}}(z_1,\ldots,z_n)
$$
where the max is over points $z_1,\ldots,z_n$, and  $\mc{N}_{\mc{A}}(z_1,\ldots,z_n)$ is the number of distinct subsets of the form 
$$
A_f \cap \{z_1, \ldots, z_n\}
$$
as $f$ ranges over $\mc{F}$. 

To conclude the proof, it suffices to show that $s(\mc{A},n) \le S(\mc{F},n)$, where the latter expression is the multiclass shatter coefficient defined above. We show this as follows.

Consider fixed pairs $z_i = (x_i, y_i)$, $i =1,\ldots, n$. Supposed that there are $N$ distinct subsets of the form $A_f \cap \{z_1, \ldots, z_n\}$, and let $f_1, \ldots, f_N$ be the classifiers in $\mc{F}$ that realize these distinct subsets. Consider the map that sends $f_i$ to the vector of its values at $x_1, \ldots, x_n$:
$$
f_i \mapsto (f_i(x_1), \ldots, f_i(x_n)) \in \mc{Y}^n.
$$
We will show that this map is injective, from which the claim follows. To see injectivity, consider classifiers $f_i$ and $f_j$, where $i \ne j$. Since $f_i$ and $f_j$ yield different subsets, it means there is some pair $(x_k, y_k)$ such that one of $f_i$ and $f_j$ classifies the pair correctly, while the other does not. This implies that $f_i(x_k) \ne f_j(x_k)$, and therefore
$$
(f_i(x_1), \ldots, f_i(x_n)) \ne (f_j(x_1), \ldots, f_j(x_n)).
$$
This concludes the proof.
\end{proof}
Now, Lemma \ref{lem:dgl:12.1} follows from the above theorem (stated in terms of the noisy data/distribution/risk) in precisely the same way that Corollary 12.1 in \citet{devroye1996probabilistic} follows from Theorem 12.6 in the same book.

\subsubsection{Proof of upper bound: \Cref{thm:risk_bound_smooth_margin}}
\label{subsec:smooth_margin_erm_proof}


\begin{theorem*}[Excess Risk Upper Bound of NI-ERM under smooth relative margin condition] 
    Let $\epsilon \in [0, 1] , \alpha > 0, C_{\alpha} > 0$. 
    Consider any $(P_X, \veta, \widetilde{\veta}) \in {\Pi}'(\epsilon, \alpha, C_{\alpha})$, assume function class $\mathcal{F}$ has Natarajan dimension $V$, and the noisy Bayes classifier $\tilde{f}^*$ belongs to $\mathcal{F}$.
    Let $\hat{f} \in \mathcal{F}$ be the ERM trained on $ Z^n = \big\{ ( X_i, \widetilde{Y}_i ) \big\}_{i=1}^n $.
    Then 
    \begin{align*}
        \expect{Z^n}{R \left( \hat{f} \right) - R\left(f^*\right)}  
        & \le \epsilon + \inf_{\kappa > 0} \left\{ C_\alpha \kappa^\alpha + \tilde{\mathcal{O}} \left( \frac{1}{\kappa} \sqrt{\frac{V}{n}} \right) \right\} \\
        & = \epsilon + \tilde{\mathcal{O}} \left( n^{-\alpha/(2 + 2\alpha)} \right).
    \end{align*}
\end{theorem*}

\begin{proof}
    Again, using \Cref{lem:excess_risk}, and \Cref{thm:erm_generalization},
    we can conclude the following, where $C$ is some large enough constant. 
    \begin{align*}
        & \mathbb{E}_{Z^n}[ R(\hat{f}) - R(f^*)] \\&\qquad\le \inf_{\kappa > 0} \left\{ P_X \Big( \mc{X} \setminus \mc{A}_{\kappa} \left(\veta, \widetilde{\veta} \right) \Big) + \frac{1}{\kappa} \left( \widetilde{R}(f) - \widetilde{R} \left( \tilde{f}^* \right) \right) \right\} \\
            &\qquad\le \inf_{\kappa > 0} \left\{ P_X \Big( \mc{X} \setminus \mc{A}_{\kappa} \left(\veta, \widetilde{\veta} \right) \Big) + \frac{1}{\kappa}  \sqrt{ \frac{CV \log(n K)}{n}}   \right\}.
    \end{align*}

    Now, by definition of $\Pi'(\epsilon, \alpha, C_{\alpha}),$ 
    it holds that
    \begin{align*}
        \forall \kappa > 0, \ P_X(\mc{M}(x;\veta,\widetilde\veta) \leq \kappa ) \le C_\alpha \kappa^\alpha + \epsilon.
    \end{align*}
    
    Thus, we can further conclude that \begin{align*}
        \mathbb{E}_{Z^n}[ R(\hat{f}) - R(f^*)] &\le \inf_{\kappa > 0} \left\{ \epsilon + C_\alpha \kappa^\alpha + \frac{1}{\kappa} \sqrt{ \frac{CV \log(nK)}{n}}\right\}.
    \end{align*} 
    The final statement now comes from optimizing the above bound, which is attained by taking the derivative w.r.t. $\kappa$ and set to zero, we have
    \begin{align*}
        \kappa_* = \left( (\alpha C_\alpha)^{-1} \sqrt{CV\log(nK)/n} \right)^{1/(\alpha + 1)}.
    \end{align*}
    This yields the bound 
    \[ \mathbb{E}_{Z^n}[ R(\hat{f}) - R(f^*)]  \le \epsilon + \mc{O} \left( \left(\sqrt{V\log(nK)/n}\right)^{\alpha/(\alpha +1)} \right) = \epsilon + \tilde{\mc{O}}(n^{-\alpha/(2\alpha + 2)}). \]
\end{proof}

\subsubsection{Proof of immunity results: Theorem \ref{thm:immunity_onehot} and \ref{thm:symmetric_noise}}
\label{subsec:immunity_proof}

Here, we state the immunity theorems in an equivalent but different way, so that the proofs are easier to follow.

\begin{theorem*}[Immunity for one-hot vector]
    Denote $\mc{B} = \{\ve_1, \dots, \ve_K \}$ to be the set of one-hot vectors.
    \begin{align*}
        & \forall \ \veta(x) \in \mc{B}, \quad \argmax{} \veta(x) = \argmax{} \E(x)^\top \veta(x) \\
        \Longleftrightarrow \quad &
        \text{Diagonal elements of $\E(x)$ maximizes its row.}
    \end{align*}
\end{theorem*}

\begin{proof}
    Let $\veta(x) = \ve_y $ for some $y$, then 
    \begin{align*}
        \widetilde{\veta}(x) = \E^\top \veta(x) =  \begin{bmatrix}
            \prob{\widetilde{Y} = 1 \mid Y = y, X = x } \\
            \prob{\widetilde{Y} = 2 \mid Y = y, X = x } \\
            \vdots \\
            \prob{\widetilde{Y} = K \mid Y = y, X = x }
        \end{bmatrix} 
        = \left[ \E(x) \right]^{^\top}_{y,:}
    \end{align*}
    To have  
    \begin{align*}
        \argmax{} \widetilde{\veta}(x) 
        = \argmax{} \left[ \E(x) \right]^{^\top}_{y,:}
        = \argmax{} \veta(x) = y 
    \end{align*}
    for any choice of $y$, it is equivalent to say that the diagonal elements of $\E(x)$ maximizes its row.
\end{proof}

\begin{theorem*}[Universal Immunity]
    Consider $K$-class classification, 
    \begin{align*}
        & \forall \ \veta(x), \quad \argmax{} \veta(x) = \argmax{} \E(x)^\top \veta(x) \\
        \Longleftrightarrow \quad &
        \exists \ e(x) \ \text{s.t. } \ \forall x, e(x) \in \Big[0, \frac{1}{K} \Big) \ \text{ and} \\
        &
        \E(x) = \begin{bmatrix}
            1 - (K-1) e(x) & e(x) & \cdots & e(x) \\ 
            e(x) & 1 - (K-1) e(x) & \cdots & e(x) \\ 
            \vdots & \vdots & \ddots & \vdots \\ 
            e(x) & e(x) & \cdots & 1 - (K-1) e(x)
        \end{bmatrix}.
    \end{align*}


\end{theorem*}

\begin{proof}

    $\Longleftarrow$: Plug $\E(x)$ into the expression
    \begin{align*}
        \widetilde{\veta}(x) & = \E^\top \veta(x) \\
        & = \begin{bmatrix}
            1 - (K-1) e(x) & e(x) & \cdots & e(x) \\ 
            e(x) & 1 - (K-1) e(x) & \cdots & e(x) \\ 
            \vdots & \vdots & \ddots & \vdots \\ 
            e(x) & e(x) & \cdots & 1 - (K-1) e(x)
        \end{bmatrix} \veta(x)
        \\
        & = \left(
            \left(1 - K \cdot e(x) \right) \cdot 
            \begin{bmatrix}
                1  & 0 & \cdots & 0 \\ 
                0 & 1 & \cdots & 0 \\ 
                \vdots & \vdots & \ddots & \vdots \\ 
                0 & 0 & \cdots & 1
            \end{bmatrix} + e(x) \cdot
            \begin{bmatrix}
                1 \\ 1 \\ \vdots \\ 1
            \end{bmatrix} \cdot 
            \begin{bmatrix}
                1 & \cdots & 1
            \end{bmatrix}
        \right)  \veta(x)
        \\
        & = \left(1 - K \cdot e(x) \right) \veta(x) + \text{ constant vector}
    \end{align*}
    When $e(x) \in [0, \frac{1}{K})$, we have 
    \begin{align*}
        \forall \veta(x), \  \argmax{} \widetilde{\veta} (x) = \argmax{} \veta(x).
    \end{align*}
    $\Longrightarrow$: 
    Denote $ \T(x) := \E(x)^\top$, then
    \begin{align*}
        \T(x) = \begin{bmatrix}
            t_{11}(x) & t_{12}(x) & \cdots & t_{1K}(x) \\
            t_{21}(x) & t_{22}(x) & \cdots & t_{2K}(x) \\
            \vdots & \vdots & \ddots & \vdots \\
            t_{K1}(x) & t_{K2}(x) & \cdots & t_{KK}(x)
        \end{bmatrix}
    \end{align*}
    has each column sum to $1$.
    Let us consider several choices of $\veta(x)$, which pose conditions on matrix $\T(x)$.

    1) If $\veta(x) =  \left[\frac{1}{K} \ \frac{1}{K} \cdots \frac{1}{K} \right]^\top$, then 
    \begin{align*}
        \widetilde{\veta}(x) = \T(x) \veta(x) = \frac{1}{K} 
        \begin{bmatrix}
            t_{11}(x) + t_{12}(x) + \cdots + t_{1K}(x) \\ 
            t_{21}(x) + t_{22}(x) + \cdots + t_{2K}(x) \\
            \vdots \\
            t_{K1}(x) + t_{K2}(x) + \cdots + t_{KK}(x) 
        \end{bmatrix}.
    \end{align*}
    To have 
    \begin{align*}
        \argmax{} \widetilde{\veta}(x) = \argmax{} \veta(x) = \left\{1,2, \dots, k \right\},
    \end{align*}
    all elements of $\widetilde{\veta}(x)$ must be equal, i.e., each row of $\T(x)$ should sum to the same value.
    The sum of all elements in $\T(x)$ is $K$, since all column sum to $1$.
    Therefore, each row of $\T(x)$ also sum to $1$.

    2) If $\veta(x) = \left[\frac{1}{K-1} \ \frac{1}{K-1} \cdots  \frac{1}{K-1} \ 0 \right]^\top $, then
    \begin{align*}
        \widetilde{\veta}(x) = \T(x) \veta(x) & = \frac{1}{K-1} 
        \begin{bmatrix}
            t_{11}(x) + t_{12}(x) + \cdots + t_{1(K-1)}(x) \\ 
            t_{21}(x) + t_{22}(x) + \cdots + t_{2(K-1)}(x) \\
            \vdots \\
            t_{(K-1)1}(x) + t_{(K-1)2}(x) + \cdots + t_{(K-1)(K-1)}(x) \\
            t_{K1}(x) + t_{K2}(x) + \cdots + t_{K(K-1)}(x) 
        \end{bmatrix} \\
        & = \frac{1}{K-1} 
        \begin{bmatrix}
            1 -  t_{1K}(x) \\ 
            1 - t_{2K}(x) \\
            \vdots \\
            1 - t_{(K-1)k}(x) \\
            1 - t_{KK}(x)
        \end{bmatrix}. \qquad \because \text{each row of $\T(x)$ sum to $1$}
    \end{align*}

    To have 
    \begin{align*}
        \argmax{} \widetilde{\veta}(x) = \argmax{} \veta(x) = \left\{1,2, \dots, K-1 \right\},
    \end{align*}
    the first $K-1$ elements of $\widetilde{\veta}(x)$ must be equal (and larger than $t_{KK}(x)$), then we have 
    \begin{align*}
        t_{1K}(x) = t_{2K}(x) = \cdots = t_{(K-1)K}(x).
    \end{align*}
    In other words, all elements of the $K$-th column of $\T(x)$ are the same (except for the $(K, K)$-th element).
    Similarly, consider $\veta(x)$ to be a vector that contains $0$ in the $i$-th position and $\frac{1}{K-1}$ in other positions, then the general condition for $\T(x)$ is that:
    all elements of the $i$-th column are equal, except the $i$-th diagonal.
    Written explicitly,
    \begin{align*}
        \T(x) = \begin{bmatrix}
            t_{11}(x) & t_{12}(x) & t_{13}(x) & \cdots & t_{1K}(x) \\
            t_{21}(x) & t_{22}(x) & t_{13}(x) & \cdots & t_{1K}(x) \\
            t_{21}(x) & t_{12}(x) & t_{33}(x) & \cdots & t_{1K}(x) \\
            \vdots & \vdots & \vdots & \ddots & \vdots \\
            t_{21}(x) & t_{12}(x) & t_{13}(x) & \cdots & t_{KK}(x)
        \end{bmatrix}.
    \end{align*}

    Since each row and column of $\T(x)$ sum to $1$, we have
    \begin{align*}
        \begin{cases}
            0 + t_{12}(x) + t_{13}(x) + \cdots + t_{1K}(x) & = \ (K-1) t_{21}(x) \longleftarrow \text{sum of first row = sum of first column} \\
            t_{21}(x) + 0 + t_{13}(x) + \cdots + t_{1K}(x) & = \ (K-1) t_{12}(x)  \\
            t_{21}(x) + t_{12}(x) + 0 + \cdots + t_{1K}(x) & = \ (K-1) t_{13}(x)  \\
            & \ \vdots \\
            t_{21}(x) + t_{12}(x) + t_{13}(x) + \cdots + 0 & = \ (K-1) t_{1K}(x)
        \end{cases}
    \end{align*}
    Subtracting the first equation from the second, we have $t_{12}(x) = t_{21}(x)$. Repeating for all pairs of equations, we have $t_{21}(x) = t_{12}(x) = t_{13}(x) = \cdots = t_{1K}(x)$. What's more, all diagonal elements of $\T(x)$ will be equal. Thus, 
    \begin{align*}
        \T(x) =
        \begin{bmatrix}
            1 - (K-1) e(x) & e(x) & \cdots & e(x) \\ 
            e(x) & 1 - (K-1) e(x) & \cdots & e(x) \\ 
            \vdots & \vdots & \ddots & \vdots \\ 
            e(x) & e(x) & \cdots & 1 - (K-1) e(x)
        \end{bmatrix}, \text{ where } e(x) \in [0, 1].
    \end{align*}

    3) The final step is to determine what value $e(x)$ can take.
    Take $\veta(x) = \ve_y$ for some $y$, then from Theorem \ref{thm:immunity_onehot}, we know that the diagonal elements of $\T(x)$ maximize their column, therefore 
    \begin{align*}
        1 - (K-1) e(x) > e(x) \quad \Longrightarrow \quad e(x) \in [0, \frac{1}{K} \big).  
    \end{align*}

    Finally, take any $\veta(x)$, the $\argmax{}$ is preserved by multiplying this specific choice of $\T(x)$. This concludes the $\Longleftarrow$ part.
\end{proof}

\subsection{Experimental details}

\subsubsection{2D Gaussian with synthetic label noise}
\label{subsec:data_simulation}

For 2D Gaussian mixture data, we draw from two Gaussian centered at $[1 \ 1]^\top$ and $[-1 \ -1]^\top$, with covariance matrix being identity, $200$ data points from each, with label $Y = 1, 2$ respectively. To generate noisy labels, we flip every label uniformly with some probability.
We use Sklearn's logistic regression (with no $\ell_2$ regularization).
The experiment was conducted on AMD Ryzen 5 3600 CPU.  The goal of the simulation is to experimentally verify noise immunity results in Section \ref{sec:immunity}. Notice that different trial corresponds to different draw of both instances and noisy labels.

\begin{table}[H]
\centering
\caption{Testing accuracy of logistic regression on gaussian mixture data with uniform label noise. ``Noise rate'' refers to $\bb{P} \big(\widetilde{Y} \neq Y \big)$, the percentage of wrong labels in the training data. As theory in Section \ref{sec:immunity} predicts, when $\bb{P} \big(\widetilde{Y} \neq Y \big)$ reach $50\%$, there is a sharp decrease in performance. }
\vspace{1em}

\resizebox{\textwidth}{!}{%
  \begin{tabular}{cccccccccccccccccccccc}
    \hline
    Noise rate & 0 & 0.05 & 0.1 & 0.15 & 0.2 & 0.25 & 0.3 & 0.35 & 0.4 & 0.45 & 0.5 & 0.55 & 0.6 & 0.65 & 0.7 & 0.75 & 0.8 & 0.85 & 0.9 & 0.95 & 1 \\ \hline
    
    Trial \#1 & 93.00 & 92.83 & 92.38 & 92.08 & 91.78 & 91.93 & 92.25 & 92.90 & 91.83 & 92.58 & 74.68 & 25.12 & 9.70 & 7.73 & 7.52 & 7.25 & 7.38 & 7.15 & 7.18 & 7.10 & 7.00 \\
    
    Trial \#2 & 91.73 & 91.60 & 92.05 & 91.63 & 91.78 & 91.78 & 91.68 & 91.63 & 91.55 & 91.48 & 80.40 & 21.10 & 9.93 & 8.55 & 8.38 & 8.22 & 8.20 & 8.35 & 8.33 & 8.40 & 8.28 \\
    
    Trial \#3 & 92.73 & 92.75 & 92.78 & 92.78 & 92.58 & 92.45 & 91.68 & 88.15 & 82.58 & 59.83 & 49.53 & 35.80 & 21.28 & 14.35 & 9.33 & 8.53 & 8.12 & 7.70 & 7.13 & 7.23 & 7.28 \\
    
    Trial \#4 & 91.55 & 91.58 & 91.60 & 91.63 & 91.68 & 91.60 & 91.25 & 90.98 & 89.98 & 86.38 & 60.53 & 9.95 & 8.75 & 10.00 & 10.45 & 9.08 & 9.00 & 9.53 & 9.20 & 9.03 & 8.45 \\
    
    Trial \#5 & 91.55 & 91.58 & 91.60 & 91.63 & 91.68 & 91.60 & 91.25 & 90.98 & 89.98 & 86.38 & 60.53 & 9.95 & 8.75 & 10.00 & 10.45 & 9.08 & 9.00 & 9.53 & 9.20 & 9.03 & 8.45 \\ \hline
    
    Mean & 92.11 & 92.07 & 92.08 & 91.95 & 91.90 & 91.87 & 91.62 & 90.93 & 89.18 & 83.33 & 65.13 & 20.40 & 11.68 & 10.10 & 9.23 & 8.43 & 8.34 & 8.45 & 8.21 & 8.16 & 7.89 \\
    
    Std & 0.70 & 0.66 & 0.51 & 0.50 & 0.38 & 0.35 & 0.41 & 1.74 & 3.79 & 13.44 & 12.35 & 10.94 & 5.39 & 2.56 & 1.29 & 0.75 & 0.68 & 1.07 & 1.03 & 0.94 & 0.70 \\
    
  \end{tabular}
}
\end{table}

\subsubsection{MNIST with synthetic label noise}
We flip the clean training label of MNIST (\url{http://yann.lecun.com/exdb/mnist/}) uniformly (to any of the wrong classes). We use a shallow neural network with two convolution layers and two fully connected layers. We train with stochastic gradient descent with learning rate $0.01$ for $10$ epochs, batch size equals $64$. We use the same hyperparamters for all tests. The experiments were conducted on a single NVIDIA GTX 1660S GPU. The goal of the simulation is to experimentally verify noise immunity results in Section \ref{sec:immunity}. Here randomness corresponds to different realization of noisy labels and stochastic gradient descent.

\begin{table}[H]
\centering
\caption{Testing accuracy of a shallow CNN (2 conv layers with 2 fully connected layers) on MNIST with uniform label noise. ``Noise rate'' refers to $\bb{P} \big(\widetilde{Y} \neq Y \big)$, the percentage of wrong labels in the training data. As theory in Section \ref{sec:immunity} predicts, when $\bb{P} \big(\widetilde{Y} \neq Y \big)$ reach $90\%$, there is a sharp decrease in performance. }
\vspace{1em}

\resizebox{\textwidth}{!}{%
  \begin{tabular}{cccccccccccccccccccccc}
    \hline
    Noise rate & 0 & 0.05 & 0.1 & 0.15 & 0.2 & 0.25 & 0.3 & 0.35 & 0.4 & 0.45 & 0.5 & 0.55 & 0.6 & 0.65 & 0.7 & 0.75 & 0.8 & 0.85 & 0.9 & 0.95 & 1 \\
    \hline
    Trial \#1 & 98.97 & 98.89 & 98.81 & 98.46 & 98.49 & 98.16 & 98.46 & 98.07 & 97.98 & 97.57 & 97.88 & 97.84 & 97.19 & 97.10 & 96.70 & 95.02 & 89.00 & 83.72 & 11.58 & 0.17 & 0.03 \\

    Trial \#2 & 98.88 & 98.73 & 98.94 & 98.55 & 98.72 & 98.66 & 98.50 & 98.24 & 98.15 & 98.23 & 97.86 & 97.98 & 97.70 & 97.10 & 96.91 & 95.76 & 91.99 & 88.49 & 9.99 & 0.08 & 0.04 \\

    Trial \#3 & 99.00 & 99.04 & 98.86 & 98.56 & 98.69 & 98.66 & 98.51 & 98.49 & 98.37 & 98.25 & 98.25 & 97.39 & 97.37 & 97.18 & 96.66 & 94.88 & 92.15 & 81.48 & 6.19 & 0.14 & 0.04 \\

    Trial \#4 & 99.04 & 98.86 & 98.70 & 98.76 & 98.83 & 98.65 & 98.34 & 98.42 & 98.58 & 98.47 & 98.00 & 97.41 & 97.63 & 97.09 & 96.46 & 95.94 & 93.19 & 84.78 & 8.68 & 0.19 & 0.01 \\

    Trial \#5 & 99.05 & 98.58 & 98.89 & 98.82 & 98.72 & 98.83 & 98.34 & 98.55 & 98.40 & 98.38 & 98.01 & 97.31 & 97.33 & 96.21 & 96.29 & 94.92 & 90.38 & 85.84 & 8.98 & 0.13 & 0.08 \\
    \hline
    Mean & 98.99 & 98.82 & 98.84 & 98.63 & 98.69 & 98.59 & 98.43 & 98.35 & 98.30 & 98.18 & 98.00 & 97.59 & 97.44 & 96.94 & 96.60 & 95.30 & 91.34 & 84.86 & 9.08 & 0.14 & 0.04 \\

    Std & 0.07 & 0.17 & 0.09 & 0.15 & 0.12 & 0.25 & 0.08 & 0.20 & 0.23 & 0.36 & 0.16 & 0.30 & 0.21 & 0.41 & 0.24 & 0.51 & 1.65 & 2.59 & 1.98 & 0.04 & 0.03 \\
  \end{tabular}
}
\end{table}

\subsubsection{CIFAR with synthetic label noise}
We flip the clean training label of CIFAR-10 (\url{https://www.cs.toronto.edu/~kriz/cifar.html}) uniformly (to any of the wrong classes). To have a fair comparison between different methods, we fix the realization of noisy labels. Follow the 2-step procedure described in Section \ref{sec:practical}, we use different pre-trained neural networks as feature extractor: forward-passing the training image through the network and record the feature. Then use sklearn's (\url{https://scikit-learn.org/stable/}) logistic regression function to fit the $(\text{feature, noisy label})$ pair in a full batch manner. We pre-specify a range of values for $\ell_2$ regularization ($\{0.0001, 0.001, 0.01, 0.1, 1, 10, 100\}$ ) and number of iterations for lbfgs optimizer ($\{10, 20, 50, 100 \}$), then do cross-validation on noisy data to pick the best hyper-parameters. We use the same range of hyper-parameters in all tests. The experiments were conducted on a single NVIDIA Tesla V100 GPU. The result is deterministic.

\begin{table}[h]
\centering
\caption{Peformance on CIFAR-10 with synthetic label noise. We apply linear model on top of different feature extractors: ``ResNet-50 TL'' refers to using a pre-trained ResNet-50 on ImageNet \citep{deng2009imagenet} (available in Pytorch model library) in a transfer learning fashion, ``ResNet-50 SSL'' refers to using a pre-trained ResNet-50 on unlabeled CIFAR data with self-supervised loss \citep{chen2020simple} (publicly downloadable weights \url{https://github.com/ContrastToDivide/C2D?tab=readme-ov-file}) and ``DINOv2 SSL'' refers to using the self-supervised foundation model DINOv2 \citep{oquab2023dinov2} (available at \url{https://github.com/facebookresearch/dinov2}) as the feature extractor. ``Noise rate'' refers to $\bb{P} \big(\widetilde{Y} \neq Y \big)$, the percentage of wrong labels in the training data. As theory in Section \ref{sec:immunity} predicts, when $\bb{P} \big(\widetilde{Y} \neq Y \big)$ reach $90\%$, there is a sharp decrease in performance. We employed Python's sklearn logistic regression and cross-validation functions without data augmentation. The results are deterministic.}
\vspace{1em}

\resizebox{\textwidth}{!}{%
  \begin{tabular}{cccccccccccccc}
    \hline
    Noise rate & 0 & 0.1 & 0.2 & 0.3 & 0.4 & 0.5 & 0.6 & 0.7 & 0.8 & 0.85 & 0.9 & 0.95 & 1 \\
    \hline
    Linear & 41.37 & 41.09 & 40.97 & 40.37 & 40.45 & 39.44 & 37.28 & 35.20 & 26.74 & 18.00 & 10.28 & 5.50 & 3.92 \\
    
    Linear + ResNet-50 TL & 90.17 & 89.58 & 89.01 & 88.27 & 87.55 & 87.28 & 86.40 & 85.01 & 82.03 & 74.02 & 10.82 & 1.47 & 0.26 \\
    
    Linear + ResNet-50 SSL & 92.48 & 92.26 & 91.74 & 91.46 & 91.13 & 90.33 & 91.07 & 90.99 & 89.11 & 83.89 & 10.08 & 1.31 & 0.34 \\
    
    Linear + DINOv2 SSL & 99.25 & 99.27 & 99.23 & 99.14 & 99.10 & 99.11 & 99.02 & 98.84 & 95.50 & 76.91 & 10.13 & 0.92 & 0.03 \\
    \hline
  \end{tabular}
}
\end{table}

\subsubsection{CIFAR with human label error}
We load the noisy human labels provided by \url{http://noisylabels.com/}, then follow exact the same procedure as above.

\begin{table}[H]
\centering
\caption{Performance on CIFAR-N dataset (\url{http://noisylabels.com/}) in terms of testing accuracy. ``Aggre'', ``Rand1'', \dots, ``Noisy'' denote various types of human label noise. We apply linear model on top of different feature extractors: ``ResNet-50 TL'' refers to using a pre-trained ResNet-50 on ImageNet \citep{deng2009imagenet} in a transfer learning fashion, ``ResNet-50 SSL'' refers to using a pre-trained ResNet-50 on unlabeled CIFAR data with self-supervised loss \citep{chen2020simple} and ``DINOv2 SSL'' refers to using the self-supervised foundation model DINOv2 \citep{oquab2023dinov2} as the feature extractor.
We employed Python's sklearn logistic regression and cross-validation functions without data augmentation; the results are deterministic and directly reproducible.}
\vspace{1em}

\resizebox{\textwidth}{!}{
\begin{tabular}{@{}ccccccc@{}}
\toprule
Methods     & \multicolumn{5}{c}{CIFAR-10N}            & CIFAR-100N \\ \cmidrule(lr){2-6} \cmidrule(l){7-7}
    & Aggre & Rand1 & Rand2 & Rand3 & Worst & Noisy \\ \midrule
Linear & 40.73 &	40.41 &	40.31 &	40.63 &	38.43 & 16.61
 \\
Linear + ResNet-50 TL & 89.18 &	88.63 &	88.61 &	88.66 &	85.32 & 62.89
 \\
Linear + ResNet-50 SSL & 91.78 & 91.66 & 91.39 & 91.28 & 87.84 & 57.95
 \\
Linear + DINOv2 SSL & 98.69  & 98.80  & 98.65  & 98.67  & 95.71  & 83.17 \\ \bottomrule
\end{tabular}
}

\end{table}


\subsection{Additional experiments}
\label{subsec:additional_experiments}

\subsubsection{Linear probing, then fine tuning (LP-FT)}
We study whether `linear probing, then fine tuning' (LP-FT) \citep{kumar2022fine} works better than linear probing (LP) only, in label noise learning scenario.


\begin{table}[H]
\centering
\caption{Performance on CIFAR-N dataset (\url{http://noisylabels.com/}) in terms of testing accuracy. ``Clean'' refers to no label noise, ``Aggre'', ``Rand1'', \dots, ``Noisy'' denote various types of human label noise. We compare the testing accuracy of LP-FT versus LP only, over different feature extractors: ``ResNet-50 TL'' refers to using a pre-trained ResNet-50 on ImageNet \citep{deng2009imagenet} in a transfer learning fashion, ``ResNet-50 SSL'' refers to using a pre-trained ResNet-50 on unlabeled CIFAR data with contrastive loss \citep{chen2020simple} and ``DINOv2 (small) SSL'' refers to using a light version of the self-supervised foundation model DINOv2 \citep{oquab2023dinov2} as the feature extractor.
}
\vspace{0.5em}

\resizebox{\textwidth}{!}{
\begin{tabular}{@{}lccccccccc@{}}
\toprule
Feature & Method     & \multicolumn{6}{c}{CIFAR-10N}            & \multicolumn{2}{c}{CIFAR-100N} \\ 
\cmidrule(lr){3-8} \cmidrule(l){9-10}
  &  & Clean & Aggre & Rand1 & Rand2 & Rand3 & Worst & Clean & Noisy \\ \midrule
\multirow{2}{*}{ResNet-50 TL} & LP (ours) & 90.17 & 89.18 &	\bf{88.63} & \bf{88.61} & \bf{88.66} & \bf{85.32} & 71.79 & 62.89
 \\ 
& LP-FT & \bf{95.94} & \bf{92.03}  & 88.55 & 87.78 & 87.82 & 71.88 & \bf{82.3} & \bf{63.85}
 \\ \hline
\multirow{2}{*}{ResNet-50 SSL} & LP (ours) & 92.54 & \bf{91.78} & \bf{91.66} & \bf{91.46} & \bf{91.17} & \bf{87.85} & 69.88 & \bf{57.98}
 \\
& LP-FT & \bf{94.11} & 89.11  & 84.49  & 83.75  & 84.15 & 65.00 & \bf{74.41} & 54.49 \\  \hline
\multirow{2}{*}{DINOv2 (small) SSL} & LP (ours) & 96.09 & \bf{94.8}  & \bf{94.39}  & \bf{94.42}  & \bf{94.35}  & \bf{91.14}  & 83.82 & \bf{72.46} \\ 
& LP-FT & \bf{98.23} & 93.29  & 88.03  & 87.27  & 86.94  & 67.42 & \bf{89.97}  & 64.81 \\
\bottomrule
\end{tabular}
}

\end{table}

\subsubsection{Robust learning strategy over DINOv2 feature}

This section examines how different robust learning strategy works over DINOv2 feature, compared with only training with cross entropy.


\begin{table}[H]
\centering
\caption{
Comparison of different noise robust methods on DINOv2 features. Training a linear classifier with cross entropy (CE) loss is the baseline. We compare it with robust losses: mean absolute error (MAE) loss \citep{ghosh2017robust, ma2022blessing}, sigmoid loss \citep{ghosh2015making}, and regularized approaches: `Early-Learning Regularization' (ELR) \citep{liu2020early}, `Sharpness Aware Minimization' (SAM) \citep{foret2021sharpnessaware}.
}
\vspace{0.5em}

\resizebox{\textwidth}{!}{
\begin{tabular}{@{}lccccccccc@{}}
\toprule
Feature & Method     & \multicolumn{6}{c}{CIFAR-10N}            & \multicolumn{2}{c}{CIFAR-100N} \\ 
\cmidrule(lr){3-8} \cmidrule(l){9-10}
  &  & Clean & Aggre & Rand1 & Rand2 & Rand3 & Worst & Clean & Noisy \\ \midrule
\multirow{4}{*}{DINOv2 SSL} & CE & 99.25 & 98.69  & 98.8  &  98.65 &    98.67  & 95.71 & \bf{92.85} & \bf{83.17} \\ 
& MAE & \bf{99.27} & \bf{99.04} & \bf{99.01} & \bf{99.09} & \bf{99.11} & 95.55 & 90.68 & 82.55 \\
& Sigmoid & 99.26 & 98.86 & 98.91 & 98.87 & 98.96 & \bf{96.66} & 92.82 & 82.03 \\
& ELR & 99.09 & 98.49 & 98.62 & 98.53 & 98.56 & 95.60 & 89.99 & 82.75 \\
& SAM & 99.09 & 97.66 & 98.47 & 98.53 & 98.47 & 95.47 & 89.97 & 82.85 \\
\bottomrule
\end{tabular}
}

\end{table}

\subsubsection{Synthetic instance-dependent label noise}


\begin{table}[H]
\centering
\caption{
We synthetically corrupt labels of CIFAR-10 according to \citet{xia2020part}, and compare our NI-ERM with the `Part-dependent matrix estimation' (PTD) method proposed in that same paper.
}
\vspace{0.5em}

\begin{tabular}{@{}cccccc@{}}
\toprule
Method $\backslash$ Noise rate & 10 $\%$ & 20 $\%$ &  30 $\%$ &  40 $\%$ &  50 $\%$ \\ \midrule 
PTD & 79.01 & 76.05 & 72.28 & 58.62 & 53.98 \\
NI-ERM (ours) & \bf{99.11} & \bf{98.94} & \bf{98.20} & \bf{93.35} & \bf{74.67} \\
\bottomrule
\end{tabular}

\end{table}


\newpage
\section*{NeurIPS Paper Checklist}

\begin{enumerate}

\item {\bf Claims}
    \item[] Question: Do the main claims made in the abstract and introduction accurately reflect the paper's contributions and scope?
    \item[] Answer: \answerYes{} 
    \item[] Justification: The claims are either supported by theory statements or by reproducible experiment results.

\item {\bf Limitations}
    \item[] Question: Does the paper discuss the limitations of the work performed by the authors?
    \item[] Answer: \answerYes{} 
    \item[] Justification: Limitations about our practical method is described.

\item {\bf Theory Assumptions and Proofs}
    \item[] Question: For each theoretical result, does the paper provide the full set of assumptions and a complete (and correct) proof?
    \item[] Answer: \answerYes{} 
    \item[] Justification: Assumptions are stated in the theorem statement. Full proofs are included in the appendix.

    \item {\bf Experimental Result Reproducibility}
    \item[] Question: Does the paper fully disclose all the information needed to reproduce the main experimental results of the paper to the extent that it affects the main claims and/or conclusions of the paper (regardless of whether the code and data are provided or not)?
    \item[] Answer: \answerYes{} 
    \item[] Justification: Important information about the experiments are in main text. Details on the experimental setup is described in the appendix.

\item {\bf Open access to data and code}
    \item[] Question: Does the paper provide open access to the data and code, with sufficient instructions to faithfully reproduce the main experimental results, as described in supplemental material?
    \item[] Answer: \answerYes{} 
    \item[] Justification: Code is provided, common benchmark datase were used, instructions are given, the result is easily reproducible.

\item {\bf Experimental Setting/Details}
    \item[] Question: Does the paper specify all the training and test details (e.g., data splits, hyperparameters, how they were chosen, type of optimizer, etc.) necessary to understand the results?
    \item[] Answer: \answerYes{} 
    \item[] Justification: See appendix and attached code.

\item {\bf Experiment Statistical Significance}
    \item[] Question: Does the paper report error bars suitably and correctly defined or other appropriate information about the statistical significance of the experiments?
    \item[] Answer: \answerYes{} 
    \item[] Justification: We have done repeated experiments for data simulation, e.g., different iid draw. For the result on CIFAR-N data challenge, our result is deterministic and thus no error bar.

\item {\bf Experiments Compute Resources}
    \item[] Question: For each experiment, does the paper provide sufficient information on the computer resources (type of compute workers, memory, time of execution) needed to reproduce the experiments?
    \item[] Answer: \answerYes{} 
    \item[] Justification: See appendix.
    
\item {\bf Code Of Ethics}
    \item[] Question: Does the research conducted in the paper conform, in every respect, with the NeurIPS Code of Ethics \url{https://neurips.cc/public/EthicsGuidelines}?
    \item[] Answer: \answerYes{} 
    \item[] Justification: The authors have read the NeurIPS Code of Ethics and confirm that this research follows the code of ethics.

\item {\bf Broader Impacts}
    \item[] Question: Does the paper discuss both potential positive societal impacts and negative societal impacts of the work performed?
    \item[] Answer: \answerNA{} 
    \item[] Justification: This is a theory-oriented paper towards better understanding of label noise problem.
    
\item {\bf Safeguards}
    \item[] Question: Does the paper describe safeguards that have been put in place for responsible release of data or models that have a high risk for misuse (e.g., pretrained language models, image generators, or scraped datasets)?
    \item[] Answer: \answerNA{} 
    \item[] Justification: The paper poses no such risks

\item {\bf Licenses for existing assets}
    \item[] Question: Are the creators or original owners of assets (e.g., code, data, models), used in the paper, properly credited and are the license and terms of use explicitly mentioned and properly respected?
    \item[] Answer: \answerYes{} 
    \item[] Justification: Citations and urls are included.

\item {\bf New Assets}
    \item[] Question: Are new assets introduced in the paper well documented and is the documentation provided alongside the assets?
    \item[] Answer: \answerNA{} 
    \item[] Justification: The paper does not release new assets.

\item {\bf Crowdsourcing and Research with Human Subjects}
    \item[] Question: For crowdsourcing experiments and research with human subjects, does the paper include the full text of instructions given to participants and screenshots, if applicable, as well as details about compensation (if any)? 
    \item[] Answer: \answerNA{} 
    \item[] Justification: The paper does not involve crowdsourcing nor research with human subjects.

\item {\bf Institutional Review Board (IRB) Approvals or Equivalent for Research with Human Subjects}
    \item[] Question: Does the paper describe potential risks incurred by study participants, whether such risks were disclosed to the subjects, and whether Institutional Review Board (IRB) approvals (or an equivalent approval/review based on the requirements of your country or institution) were obtained?
    \item[] Answer: \answerNA{} 
    \item[] Justification: The paper does not involve with this matter.

\end{enumerate}


\end{document}